\documentclass[lettersize,journal]{IEEEtran}
\usepackage[caption=false,font=normalsize,labelfont=sf,textfont=sf]{subfig}
\usepackage{stfloats}
\usepackage{cite}
\newcommand{\fref}[1]{Figure \ref{#1}}
\newcommand{\sref}[1]{Section \ref{#1}}
\newcommand{\tref}[1]{Table \ref{#1}}
\newcommand{\eref}[1]{Eq. (\ref{#1})}

\newcommand{\alref}[1]{Algorithm \ref{#1}}
\usepackage{graphicx}
\usepackage{amsmath}
\usepackage{amsthm}

\usepackage{amssymb,amsfonts}
\usepackage{algorithmicx}
\usepackage{array}
\usepackage{multirow}
\usepackage{mathrsfs}%
\usepackage{xcolor}%
\usepackage{textcomp}%
\usepackage{manyfoot}%
\usepackage{booktabs}%
\usepackage{algorithm}%
\usepackage{algpseudocode}%
\usepackage{listings}%
\usepackage{hyperref}
\usepackage{balance}
\usepackage{tabularx}
\usepackage{enumitem}
\setlist[enumerate]{label={(\arabic*)}}
\usepackage{soul}
\usepackage{array}
\usepackage{stfloats}
\usepackage{url}
\usepackage{verbatim}
\usepackage{epsfig}
\usepackage{color}
\usepackage{colortbl}
\usepackage{lineno}
\usepackage{graphics}
\usepackage{lmodern}
\graphicspath{{figures/}}
\usepackage{rotating}
\usepackage{float}
\usepackage{cases}
\usepackage{supertabular}
\usepackage{makecell}
\usepackage{url}

\begin{document}

\title{Deep Reinforcement Learning-Assisted Component Auto-Configuration of Differential Evolution Algorithm for Constrained Optimization: A Foundation Model}

\author{Xu~Yang,
        Rui~Wang,
        Kaiwen~Li,
        Wenhua~Li,
        and~Ling~Wang,~\IEEEmembership{Member,~IEEE,}
\thanks{Xu Yang, Rui Wang, Kaiwen Li and Wenhua Li are with the College of Systems Engineering, National University of Defense Technology, Changsha 410073, China (e-mail: yangxu616@nudt.edu.cn; rui\_wang@nudt.edu.cn; likaiwen@nudt.edu.cn, liwenhua@nudt.edu.cn).}
\thanks{Ling Wang is with the Department of Automation, Tsinghua University, Beijing 100084, China (e-mail: wangling@tsinghua.edu.cn).}
\thanks{Manuscript received September 7, 2025 (\textit{Corresponding author: Rui Wang and Kaiwen Li})}}
\markboth{Journal of \LaTeX\ Class Files,~Vol.~14, No.~8, August~2025}%
{Xu Yang \MakeLowercase{\textit{et al.}}: A Sample Article Using IEEEtran.cls for IEEE Journals}


\maketitle

\begin{center}
    \copyright\ 2025 IEEE. Personal use of this material is permitted. Permission from IEEE must be obtained for all other uses, in any current or future media, including reprinting/republishing this material for advertising or promotional purposes, creating new collective works, for resale or redistribution to servers or lists, or reuse of any copyrighted component of this work in other works.
\end{center}

\begin{abstract}
  Despite significant efforts to manually design high-performance evolutionary algorithms, their adaptability remains limited due to the dynamic and ever-evolving nature of real-world problems. The "no free lunch" theorem highlights that no single algorithm performs optimally across all problems. While online adaptation methods have been proposed, they often suffer from inefficiency, weak convergence, and limited generalization on constrained optimization problems (COPs).
  To address these challenges, we introduce a novel framework for automated component configuration in Differential Evolution (DE) algorithm to address COPs, powered by Deep Reinforcement Learning (DRL). Specifically, we propose SuperDE, a foundation model that dynamically configures DE's evolutionary components based on real-time evolution. Trained offline through meta-learning across a wide variety of COPs, SuperDE is capable of recommending optimal per-generation configurations for unseen problems in a zero-shot manner. Utilizing a Double Deep Q-Network (DDQN), SuperDE adapts its configuration strategies in response to the evolving population states during optimization. Experimental results demonstrate that SuperDE significantly outperforms existing state-of-the-art algorithms on benchmark test suites, achieving superior generalization and optimization performance.
\end{abstract}

\begin{IEEEkeywords}
constrained optimization, differential evolution, deep reinforcement learning, double deep Q-network
\end{IEEEkeywords}

\section{Introduction}
\IEEEPARstart{T}{he} pursuit of optimal solutions under constraints constitutes a fundamental challenge including engineering and science \cite{beitz1996engineering,yu2024knowledge,liu2023path}, economics and management \cite{kim2025review,gunjan2023brief,juang2007adaptive}. Without loss of generality, a constrained optimization problem (COP) can be defined as follows: Given an objective function $f: \mathbb{R}^n \rightarrow \mathbb{R}$, and a set of constraints, find a vector $\mathbf{x} \in \mathbb{R}^n$ that minimizes the objective function subject to the constraints. 
\begin{equation}
\begin{aligned}
& \text{minimize} \quad f(\mathbf{x}) \\
& \text{subject to} \quad \begin{aligned}
  & g_i(\mathbf{x}) \leq 0, \quad i = 1, \ldots, m, \\
  & h_j(\mathbf{x}) = 0, \quad j = 1, \ldots, p
\end{aligned}
\end{aligned}
\end{equation}
where $g_i(\mathbf{x})$ are the inequality constraints and $h_j(\mathbf{x})$ are the equality constraints. Constraint violation (CV) is defined to quantify the degree to which a solution violates constraints, typically calculated as

\begin{equation}
  cv(\mathbf{x}) = \sum_{i=1}^{m}max(0,g_i(\mathbf{x})) + \sum_{j=1}^{p} max(0,|h_j(\mathbf{x})| - \delta)
  \label{eqcv}
\end{equation}
where $\delta$ represents the tolerance parameter of the equality constraints. A solution $\mathbf{x}$ is considered feasible when $cv(x)= 0$; otherwise, it is deemed infeasible.

Evolutionary algorithms (EAs) are widely used for tackling COPs due to their robust search capabilities and simplicity. Comprehensive reviews of evolutionary constrained optimization have been presented by Coello et al. \cite{coello2022constraint}, Rahimi et al. \cite{rahimi2023review}, and Liang et al. \cite{liang2022survey}. Besides, Li et al. \cite{li2018two} proposed a parameter-free two-archive EA for constrained multiobjective optimization problems, where one archive prioritizes convergence and the other diversity. Mohamed et al. \cite{mohamed2018novel} introduced a novel Differential Evolution (DE) algorithm for COPs, featuring a triangular mutation rule based on convex combinations of triplet vectors and difference vectors among best, better, and worst individuals. Zhu et al. \cite{zhu2020constrained} developed a detect-and-escape strategy that used feasibility and CV changing rates to detect stagnation and adjusted violation weights for guidance. Wang et al. \cite{wang2018random} utilized random forests and radial basis function networks as surrogates to approximate objective and constraint functions in expensive data-driven constrained multiobjective combinatorial optimization. Ming et al. \cite{ming2021dual} proposed a cooperative coevolutionary algorithm with two collaborative populations.

However, real-world optimization problems are inherently dynamic and diverse, posing significant challenges for these EAs. The fixed core components in these methods, including selection strategies, evolutionary operators, and constraint handling techniques (CHTs), severely limit their adaptability \cite{ming2024constrained}. The "no free lunch" theorem underscores that no single algorithm can universally outperform others across all problem types, making it difficult to achieve consistent optimal performance in varied settings. While recent advancements have introduced adaptive strategies \cite{wu2019ensemble, wang2020adaboost, li2022self, xue2022multi, deng2022adaptive} and online adaptation models \cite{tian2023drlemo, luo2025deep} that modify components during the optimization process, these approaches typically rely on handcrafted rules or perform operator selection and tuning on a per-problem basis during each evolutionary generation. As a result, a more flexible and efficient approach is needed to enable EAs to achieve optimal performance in diverse, challenging, and constrained environments.

To this end, we present a foundation model SuperDE that leverages Deep Reinforcement Learning (DRL) to automate the configuration of DE components, thereby enhancing its ability to solve COPs. Specifically, SuperDE is trained offline through meta-learning on diverse COPs, enabling it to recommend optimal configurations for unseen problems in a zero-shot manner. By utilizing a Double Deep Q-Network (DDQN), SuperDE autonomously adjusts DE components, including mutation strategies (MSs) and CHTs during evolution. Such a model would reduce the computational burden of traditional online tuning to form a hybrid offline-online configuration paradigm. The key contributions are as follows:
\begin{itemize} 
  \item A DRL-assisted foundation momdel for component auto-configuration is introduced to automated constrained optimization algorithm configuration. Trained on a large corpus of COPs, this model recommends per-generation operator selections for unseen problems without retraining and can be fine-tuned online for further instance-specific adaptation. 
  \item A DDQN is off-trained to learn the policy that estimates the Q-value of actions under current population states, modeling the mapping between population states and algorithm components. 
  \item Population state features are designed to reflect the distribution of decision variables, objectives, and constraints. A candidate component pool- comprising 4 MSs and 7 CHTs - is employed to balance diversity and convergence while leveraging both feasible and infeasible solutions. 
\end{itemize}

The remainder of this work is structured as follows. \sref{sec_pre} reviews the related work, \sref{sec_method} details the proposed methods, \sref{sec_exp} presents the experiments and analysis, and finally, \sref{sec_con} concludes the study and outlines future research directions.

\section{Preliminary and Related work}
\label{sec_pre}
\subsection{Deep reinforcement learning}
DRL integrates deep learning and reinforcement learning to optimize policies through agent-environment interaction \cite{8103164}. It has been applied to autonomous driving \cite{9351818}, robotic control \cite{10563996}, and communication \cite{8714026, 9086766}.

DRL constructs state-action mappings via neural networks based on the Markov decision process (MDP) framework of reinforcement learning. An MDP is formally defined by the tuple $(\mathcal{S}, \mathcal{A}, \mathcal{P}, R, \gamma)$, where:
\begin{itemize}
    \item $\mathcal{S}$: Set of environment states,
    \item $\mathcal{A}$: Set of agent actions,
    \item $\mathcal{P}$: Transition probability function $P(s'|s,a)$ describing the probability of transitioning to state $s'$ from state $s$ after action $a$,
    \item $R$: Reward function $R(s,a,s')$ assigning a numerical reward to state-action-state transitions,
    \item $\gamma \in [0,1)$: Discount factor for future rewards.
\end{itemize}

Generally, the objective of training an agent is to learn a policy $\pi(a|s)$ that maps states to actions, in order to maximize the expected cumulative reward:
\begin{equation}
  J(\pi) = \mathbb{E}\left[\sum_{k=0}^{\infty} \gamma^k R_{t+k+1} \mid \pi, s_t\right]
\end{equation}
where $s_t \in S$ is the state at time $t$, and $R_{t+k+1}$ denotes the reward received at time $t+k+1$.


\subsection{Traditional adaptive algorithms}
Balancing convergence and feasibility remains a critical challenge in solving COPs. 
To address this, researchers have developed adaptive EAs.
Qiao et al. \cite{qiao2023self} proposed an evolutionary multi-task optimization algorithm DBEMTO, which employed a self-adaptive mechanism based on strategy success rates to dynamically adjust the utilization of three evolutionary strategies. Gu et al. \cite{gu2023constrained} adaptively adjusted the CV degree of infeasible solutions by incorporating K-nearest neighbor information, with population updates guided by the adjusted CV principle. Yang et al. \cite{yang2023dual} introduced a novel dominance relationship and an adaptive constraint strength strategy. The latter was designed using CV information of infeasible solutions, following a sigmoid function:
\begin{equation}
  \eta^{(t)}=E_{\text {total }}\left(\frac{2}{1+e^{-\left(\frac{T-t}{cp}\right)}}-1\right)
\end{equation}
where $E_{\text {total }}$ denotes the maximum CV in the initial population, $T$ is the termination generation for constraint strength adjustment, $t$ is the current generation, and $cp$ is a custom parameter.

In earlier work, Hamida and Schoenauer \cite{hamida2002aschea} proposed the Adaptive Segregational Constraint Handling EA (ASCHEA), which employs a population-level adaptive penalty function as its CHT. In each generation, penalty coefficients are adjusted based on the proportion of constraint-satisfying individuals: if this proportion falls below a target threshold, coefficients increase to penalize violations more severely. The adjustment formula is:
\begin{equation}
  a_j(t+1) = a_j(t) \times \text{fact} \times \left( \frac{T_{\text{target}}}{r_t(j)} \right)
\end{equation}
where $a_j(t)$ is the penalty coefficient of the $j$-th th constraint in the $t$ generation, $\text {fact}$ is a constant greater than 1, $T_{\text {target}}$ is the target proportion, and $r_t (j)$ is the proportion of individuals who satisfy the $j$-th constraint in the $t$ generation.
Peng and Deng \cite{hu2010dynamic} put forward dynamic neighborhood hybrid particle swarm optimization (DNH\_PSO). Similarly, adaptive penalty function constraint handling mechanism was introduced in DNH\_PSO. The fitness was defined as $fitness(x) = f(x) + 10^{\alpha(1-b-\rho)}*CV(x)$, where $\alpha$ was a constant parameter between 1 to 10, $\rho$ denoted as a feasible solution and infeasible solution ratio, $b$ is pre-set the reciprocal of the maximum of proportion of feasible solutions and infeasible solutions. 
Takahama and Sakai \cite{takahama2010efficient} presented $\epsilon$ADE, which adopted $\epsilon$ level and adaptive control of algorithm parameters in DE. 

However, adaptability in these studies relies on manually defined rules, introducing subjectivity and empiricism. More intelligent methods are needed to extract deeper insights into algorithm dynamics.

\subsection{DRL-based adaptive algorithms}
With the advancement of artificial intelligence, DRL has demonstrated applicability across diverse domains, including autonomous driving \cite{kiran2021deep}, radar detection and tracking \cite{thornton2020deep}, and mobile robot navigation \cite{zhu2021deep}. It has also been leveraged to enhance EAs' performance. For instance, Luo et al. \cite{luo2025deep} proposed a DRL-guided coevolutionary algorithm, which employed two populations to optimize the original COP and its unconstrained variants, respectively. DRL integrated dual evaluation metrics to quantify population, thereby effectively steering the coevolutionary process. Ming et al. \cite{ming2024constrained} framed the population state as the state space, candidate operators as actions, and population state improvements as reward signals. DRL was used to learn a policy that selected optimal operators for given population states. Recognizing that CHTs exhibit problem-specific applicability and may vary across optimization stages, Wang et al. \cite{wang2024adaptive} introduced a DRL-based adaptive CHT selection framework. This framework tailors CHT selection to distinct evolutionary states using a deep Q-learning network to evaluate the impact of CHTs and operators on population dynamics. 

However, these DRL-based automated configuration methods perform operator selection and tuning on a per-problem, per-generation basis, yielding configurations optimized only for specific problems. Concurrently, online training imposes substantial computational demands and risks bias due to limited sample sizes. Research into DRL-based adaptive DE algorithms remains nascent, with critical gaps - most notably the absence of a foundational model for automated algorithm configuration.

\section{Method: SuperDE}
\label{sec_method}
This section first introduces the overall framework, followed by detailed technical descriptions.

\subsection{Overall SuperDE}
According to \sref{secACC}, the environment includes an algorithm component candidate pool, a problem set, and a population for evolutionary optimization. To formulate the MDP, $state$, $action$, and $reward$ need to be well-designed preliminarily, introduced in the following subsections. The pseudocode of SuperDE is outlined in \alref{algDRLMCDE}. 

\begin{figure*}
  \centering
  \includegraphics[width = 0.8\textwidth]{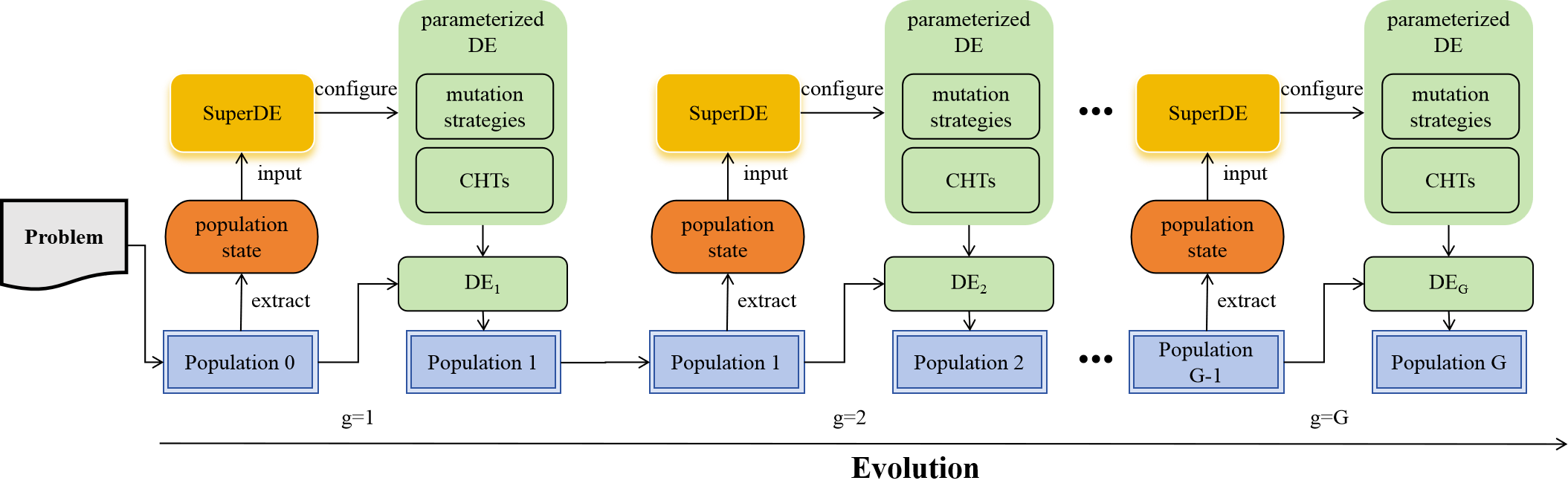}
  \caption{Overall structure of SuperDE}
  \label{figOverall}
\end{figure*}

SuperDE integrates DRL for the component auto-configuration of a parameterized algorithm $\Lambda$ using evolutionary components from the candidate pool $\Psi = \{\psi_1, \psi_2, \ldots, \psi_L\}$. The training of SuperDE begins by initializing the policy network $\pi_\theta$ and critic network $\tau_\phi$ via $initializeNetwork()$ if they do not already exist. It then iterates for a maximum of $Epo$ epochs, with each epoch involving processing $M$ instances (i.e., $M$ episodes) from the instance set $I = \{I_1, I_2, \ldots, I_M\}$. For each instance $I_j$, a population $P$ is first initialized, and a loop is executed to perform evolutionary iterations until the termination condition $(done = 1)$ is met. In each evolutionary generation $g$, the current state $s_g$ is calculated based on the population $P$, and an action $a_g$ is sampled from the policy $\pi_\theta(a_g|s_g)$ conditioned on $s_g$. This action is decoded to select a component $\psi_g$ from $\Psi$, which includes a MS $ms_g = \psi_{g,1}$ and a CHT $cht_g = \psi_{g,2}$. Using these selected components, offspring $O$ are generated from $P$ via $generateOff()$, and the new population $P'$ is obtained through environmental selection with $cht_g$. The reward $r_g$ is computed by comparing $P$ and $P'$, while the next state $s_{g+1}$ is derived from $P'$. The trajectory tuple $(s_g, a_g, r_g, s_{g+1}, done)$ is stored in the Experience Buffer $EB$, and the population is updated to $P'$ for the next step. The overall structure is presented in \fref{figOverall}. 

\begin{figure*}
  \centering
  \includegraphics[width = 0.8\textwidth]{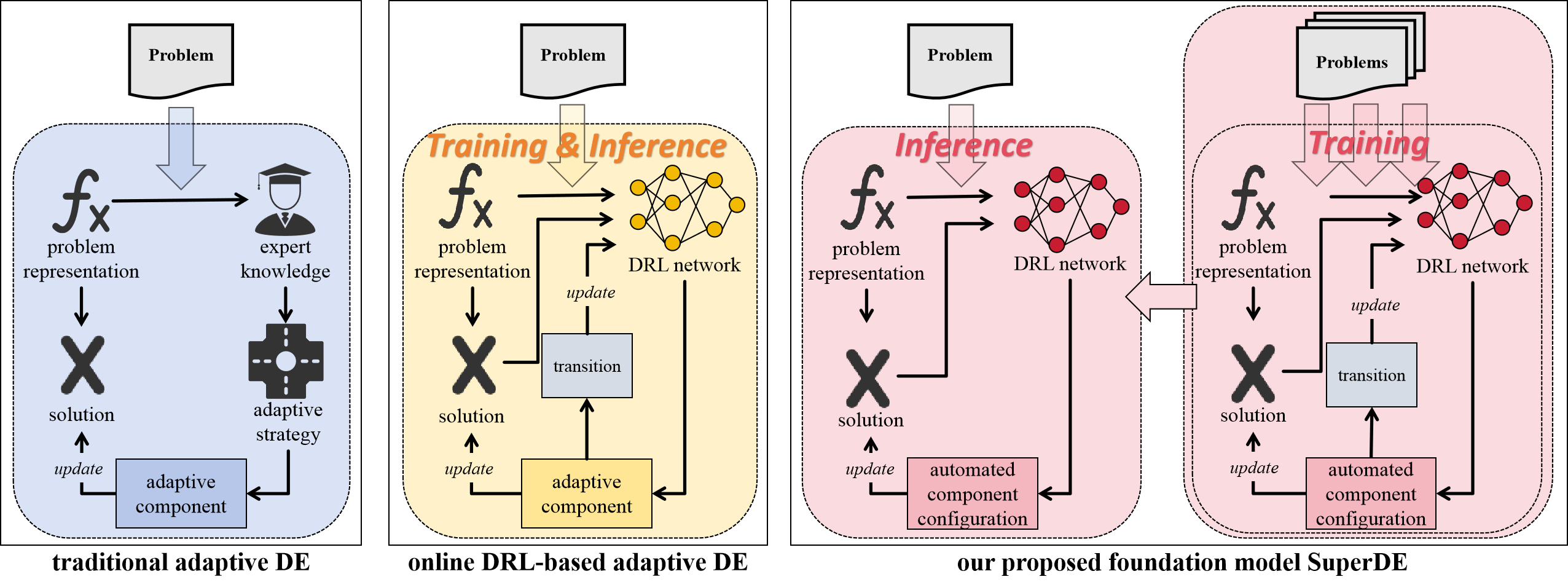}
  \caption{The schematic diagrams of traditional adaptive DE, online DRL-based adaptive DE, and our proposed foundation model SuperDE. SuperDE additionally trains the configuration policy across problems, and applies the well-trained policy to configure DE during inference.}
  \label{figSche}
\end{figure*}

In order to reduce resource consumption in online training and enhancing the generalizability, MetaBBO is introduced to train the SuperDE through a bi-level optimization framework. In the outerloop, various problems are sampled for one training epoch. In the inner loop, the DRL networks are updated using the transitions stored in $EB$ obatined in each episode (i.e., optimizing an instance $I_j$). This process repeats across all instances and epochs to optimize the policy $\pi_\theta$, which ultimately enables effective automated configuration of the EA components. The schematic diagrams of traditional adaptive DE, online DRL-based adaptive DE and our proposed SuperDE are presented in \fref{figSche}.

\begin{algorithm} \scriptsize
  \caption{Pseudocode of SuperDE}
  \begin{algorithmic}[1]
    \Require Instance set $I = \{I_1,I_2,...,I_M\}$, parameterized algorithm $\Lambda$ with candidate component pool $ \Psi = \{\psi_1, \psi_2,...,\psi_L\}$, maximum epoch number $Epo$, Experience Buffer $EB$
    \Ensure Policy $\pi_{\theta}$
    \If{not exist Policy $\pi_\theta$}
      \State $\pi_\theta \gets initializeNetwork()$
    \EndIf
    \State $epo \gets 1$
    \While {$epo \leq Epo$} 
      \State $j \gets 1$
      \While {$j \leq M$} \%$M$ episode
        \State $g \gets 1$
        \State Population $P \gets initializePopulation(I_j)$
        \State $done = 0$
        \If {$done == 0$}
          \State State $s_g \gets calculateState(P)$
          \State Action $a_g \gets \pi_\theta(a_g|s_g)$
          \State Component $\psi_g \gets decodeAction(a_g)$
          \State Mutation strategy $ms_g \gets \psi_{g,1}$
          \State Constraint handling technique $cht_g \gets \psi_{g,2}$
          \State Offspring $O \gets generateOff(P, ms_g, I_j)$
          \State $P' \gets envSelection(P,O,cht_g)$
          \State Reward $r_g \gets calculateReward(P,P')$
          \State NextState $s_{g+1} \gets calculateState(P')$
          \State $done \gets algorithmTermination(I_j)$
          \State $EB \gets storeTraj(s_g, a_g, r_g, s_{g+1}, done)$
          \State $P \gets P'$
          \State $g \gets g+1$
        \EndIf
        \State $\pi_\theta \gets updateDRL(EB)$
        \State $j \gets j+1$
      \EndWhile
      \State $epo \gets epo + 1$
    \EndWhile
  \end{algorithmic}
  \label{algDRLMCDE}
\end{algorithm}  

\subsection{Component auto-configuration formulation}
\label{secACC}
Given an algorithm $\varLambda$ and a set of problem instances $I = \{I_1,I_2,...I_j,...,I_M\}$, the objective of traditional component configuration is to find an optimal policy $\pi^*$ for a specific problem instance $I_j$ as follows:
\begin{equation}
  \pi^*=argmax_{\pi \in \Pi} \sum_{g=1}^{G} \mathcal{M}(\pi(I_j, g),I_j)
\end{equation}
where $\pi:I_j \times G \rightarrow \varLambda$ is a configurator that configures $\varLambda$ with candidate component $\psi_i \in \Psi = \{\psi_1, \psi_2,...,\psi_L\}$ for instance $I_j$ in each iteration $g$ to maximize the algorithm performance metric $\mathcal{M}: \Lambda \times I_j \rightarrow \mathcal{R}$, such as Luo et al. \cite{luo2025deep}. $G$ represents the maximum evolutionary iterations.

Consequently, traditional automated configuration methods perform operator selection and tuning on a per-problem basis during every evolutionary generation, yielding an optimal configuration specific to that problem alone. 

Pertaining to our component auto-configuration combined with MetaBBO, the objective becomes to find a comprehensively optimal policy $\pi^*$ for all given problem instances as follows:
\begin{equation}
  \pi^* = argmax_{\pi \in \Pi} \sum_{j=1}^{M} \sum_{g=1}^{G} \mathcal{M}(\pi(I_j, g),I(j))
\end{equation}

\subsection{MDP modeling}
\label{secMdpComponents}
To enable the policy network \(\pi_\theta\) to adaptively configure evolutionary components during the optimization process, we formulate the problem as a MDP with well-defined state, action, and reward components. This section details the design of each MDP element, as implemented in the proposed framework.

\subsubsection{State space design}
The state is a high-dimensional feature vector encoding the current population's characteristics, designed to capture the optimization progress, the distribution of decision variables and objectives. It includes the following features:

\begin{itemize}
  \item $feature_1$: Feasible solution ratio ($fsr$), proportion of population with zero CV, calculated as \(fsr = \frac{\sum(fs)}{N}\).
  \item $feature_{2,3,4,5}$: Best, worst, mean, and median fitness values of the current population.
  \item $feature_{6,7,8,9}$: Objectives statistics values, including minimum, maximum, mean, and median objective values of the current population.
  \item $feature_{10,11,12,13}$: CV statistics values, including minimum, maximum, mean, and median CV values.
  \item $feature_{14,15}$: Feasible region structure. $feature_{14}$: normalized number of feasible clusters (identified via DBSCAN on feasible solutions); $feature_{15}$: ratio of feasible boundary crossings.
  \item $feature_{16}$: Spearman correlation coefficient between objective values and CVs.
  \item $feature_{17,18}$: Proportions of solutions in ideal zones, measuring convergence toward high-quality regions.
  \item $feature_{19}$: Mean standard deviation of decision variables.
  \item $feature_{20}$: Mean distance from each solution to the best solution's decision variables.
  \item $feature_{21}$: Average proportion of active constraints (those violated by solutions).
  \item $feature_{22}$: Number of function evaluations consumed.
\end{itemize}

This comprehensive state representation ensures the policy network has sufficient information to assess the current optimization stage and select appropriate components.

\subsubsection{Action space design}
Prior research \cite{tian2023principled} has demonstrated that the effectiveness of algorithmic components, including CHTs \cite{liang2022survey}, can vary significantly across different optimization problems. Consequently, the action space is constructed to encode the selection of two critical evolutionary components in COPs: MS and CHT. It is designed as a discrete, two-dimensional vector \(a = [ms_{\text{num}}, cht_{\text{num}}]\), where:
\begin{itemize}
  \item \(ms_{\text{num}} \in \{1, 2, 3, 4\}\): An integer index specifying one of four pre-defined MSs from the candidate pool \(\Psi\), to generate offspring.
  \item \(cht_{\text{num}} \in \{1, 2, ..., 7\}\): An integer index selecting one of seven CHTs, to perform environmental selection between the parent $P$ offspring $O$.
\end{itemize}

The action space is formally defined as a finite set \(\mathcal{A} = \{[ms_{\text{num}}, cht_{\text{num}}] \mid ms_{\text{num}} \in \{1,...,4\}, cht_{\text{num}} \in \{1,...,7\}\}\), resulting in 28 possible action combinations. This discrete design ensures the policy network can efficiently learn a mapping from states to component selections.

\subsubsection{Reward design}
The reward \(r_g\) is designed to guide the policy toward improving both feasibility and solution quality, adapting to the optimization phase (feasible vs. infeasible). When feasible solutions exist:
\begin{equation}
  r_g = 0.4 \cdot \frac{\text{bestFitness}_{\text{prev}} - \text{bestFitness}_{\text{new}}}{|\text{bestFitness}_{\text{prev}}| + 10^{-8}} + 0.6 \cdot fsr_{\text{new}}
\end{equation}
This rewards improvements in the best feasible fitness (40\% weight) and increases in the feasible solution ratio (60\% weight), prioritizing both quality and feasibility expansion.

When no feasible solutions exist (\(fsr = 0\)):
\begin{equation}
  r_g = 0.7 \cdot (\text{avgCV}_{\text{prev}} - \text{avgCV}_{\text{new}}) + 0.3 \cdot \frac{\text{minCV}_{\text{prev}}}{\text{minCV}_{\text{new}} + 10^{-8}}
\end{equation}
where \(\text{avgCV}\) is the average CV, and \(\text{minCV}\) is the minimum CV. The reward emphasizes reducing average violations (70\% weight) and improving the best (smallest) violation (30\% weight), guiding the search toward feasibility.

This adaptive reward function aligns with the dual goals of evolutionary optimization: first achieving feasibility, then improving solution quality within the feasible region.

\subsection{DRL Agent Network for Policy Learning}
The SuperDE framework employs a DDQN agent to learn the policy \(\pi_\theta(a|s)\), which dynamically selects evolutionary components. The agent leverages a Q-value function approximator to map states to action values, with a dual-network architecture to stabilize learning. This section presents the network architecture, update rules, and operational workflow in detail.

\subsubsection{Network architecture}
The core of the agent is a Q-value network (critic) that estimates the state-action value function \(Q(s,a;\theta)\), where \(\theta\) denotes the network parameters. The architecture consists of: (1) Input processing layers: For vector observations (e.g., the 22-dimensional state in SuperDE), fully connected (fc) layers with 256 neurons per layer are used to extract features, followed by ReLU activation and flattening.
(2) Fusion and output layers: Processed features from input layers are concatenated, passed through additional fc layers (with ReLU activation), and finally mapped to a vector of Q-values. Formally, the output is \([Q(s,a_1;\theta), Q(s,a_2;\theta), ..., Q(s,a_{28};\theta)]\), where each \(Q(s,a_i;\theta)\) represents the expected cumulative reward of taking action \(a_i\) in state $s$.

To stabilize training, a target critic network with identical architecture but frozen parameters \(\theta^-\) is maintained. This target network generates delayed estimates of Q-values, reducing the correlation between target and predicted values during updates.

\subsubsection{Network update}
The agent learns via temporal difference learning, minimizing the discrepancy between predicted Q-values and target Q-values. For a transition \((s_g, a_g, r_g, s_{g+1}, {done})\) sampled from the experience buffer $EB$, the target Q-value \(y_g\) is computed using the target network. In DDQN, the action selection for the next state \(s_{g+1}\) is determined by the current critic network (to avoid overestimation), while the value is estimated by the target network:\(y_g = r_g + (1 - {done}) \cdot \gamma^k \cdot Q(s_{g+1}, a^*; \theta^-)\),where \(a^* = \arg\max_a Q(s_{g+1}, a; \theta)\) (action selected by the current critic), \(\gamma\) is the discount factor, $k$ is the look-ahead steps, and $done$ is a binary indicator for episode termination. 

The critic network parameters $\theta$ are updated to minimize the mean squared error between predicted Q-values and target Q-values, weighted by importance sampling weights from the experience buffer based on \eref{eqThetaUpdate}:
\begin{equation}
  \mathcal{L}(\theta) = \mathbb{E}_{(s,a,r,s',\text{done}) \sim {EB}} \left[ w \cdot \left( Q(s,a;\theta) - y \right)^2 \right]
  \label{eqThetaUpdate}
\end{equation}
where $w$ denotes the sample weights (from prioritized experience replay). 

The target network parameters \(\theta^-\) are periodically synchronized with the critic network to propagate learned features. This is done either every target update frequency steps according to \eref{eqTheta2Update}:
\begin{equation}
  \theta^- \leftarrow (1 - \xi) \cdot \theta^- + \xi \cdot \theta
  \label{eqTheta2Update}
\end{equation}
where \(\xi\) is the target smooth factor, balancing stability (small \(\xi\)) and adaptability (large \(\xi\)).

\subsubsection{Workflow}
The agent operates in a closed loop of exploration, experience collection, and network updates, aligned with the MDP defined in \sref{secMdpComponents}. 

First, the critic \(Q(s,a;\theta)\) and target critic \(Q(s,a;\theta^-)\) are initialized with random parameters. The experience buffer ($EB$) is initialized to store transitions, with a maximum capacity of $ExperienceBufferLength = 20000$. The exploration policy is an \(\epsilon\)-greedy strategy, which selects the action with the highest Q-value most of the time but explores random actions with probability \(\epsilon\). 

Next, for each instance \(I_j\) and generation $g$, the agent observes state \(s_g\), selects action \(a_g \sim \pi_\theta(a|s_g)\) (via the critic's Q-value estimates), and stores the transition \((s_g, a_g, r_g, s_{g+1}, done)\) in $EB$ (line 22 of Algorithm \alref{algDRLMCDE}).

Once EB contains enough samples (exceeding number of warm-start steps), the agent samples mini-batches of size $MiniBatchSize = 128$ from $EB$. For each batch, the critic network is updated via gradient descent using the loss function. The target network is updated every target update frequency steps to reflect the latest critic parameters. 

\begin{figure*}
  \centering
  \includegraphics[width = 0.7\textwidth]{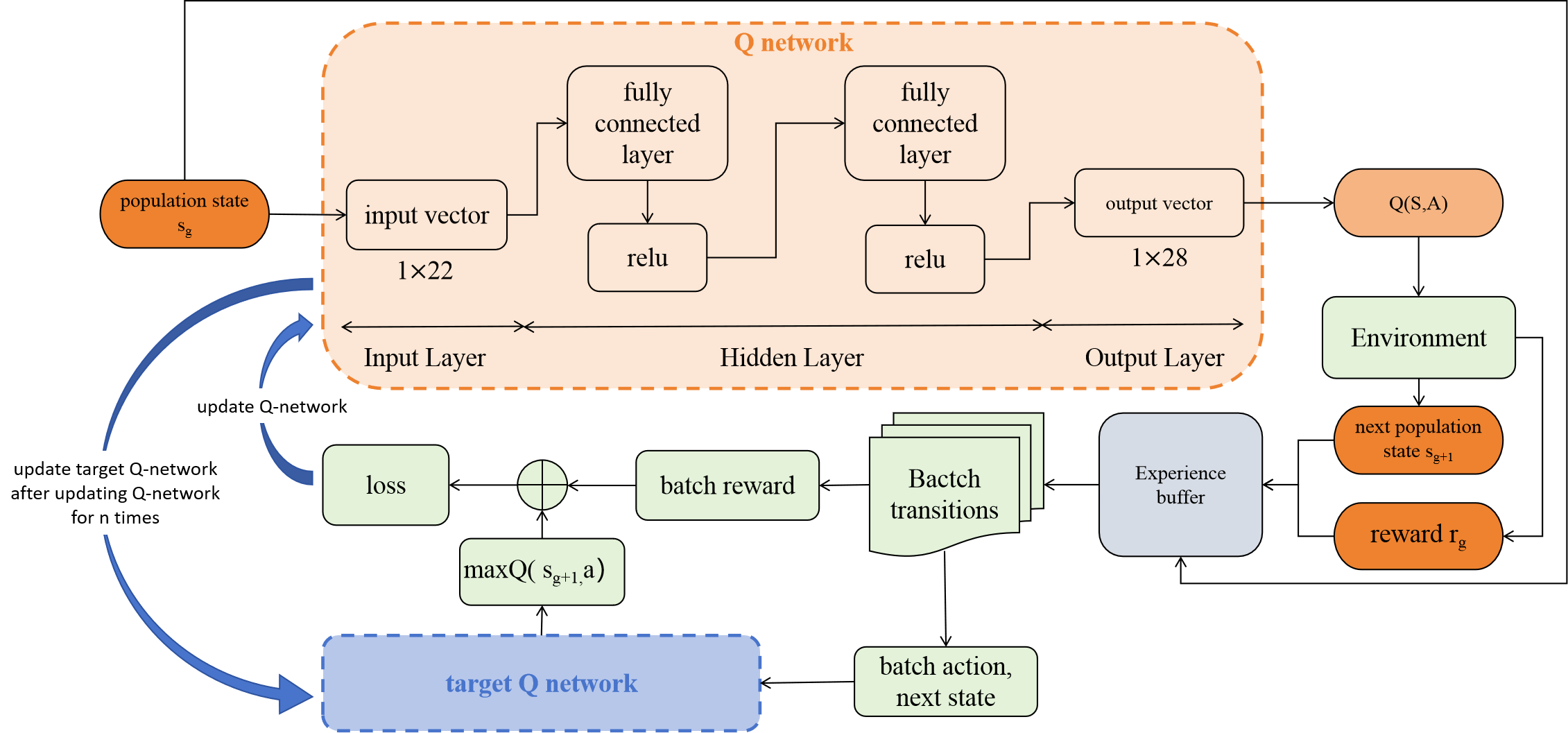}
  \caption{The workflow of DDQN}
  \label{figddqn}
\end{figure*}

The comprehensive workflow is provided in \fref{figddqn}. This dual-network architecture and experience replay mechanism ensure stable learning of the policy \(\pi_\theta\), enabling the agent to adaptively select MSs and CHTs based on real-time population states.

\subsection{Candidate pool}
The candidate pool consists of two main components, MS and CHT. 
\subsubsection{Candidate mutation strategies}
MS define how offspring are generated by perturbing parent solutions, a core step in EA. The specific four strategies, selected via \(ms_{\text{num}}\), are as follows:
\paragraph{ $MSnum = 1$ (DE/rand/1)} This strategy generates offspring by perturbing a random parent with the difference between two other random parents. Formally, for each dimension $d$ where the mask \(Site(d) = 1\) (i.e., \(rand(N,D) < CR\)), the offspring is computed as 
\begin{equation}
  O(Site) = p1(Site) + F \times (p2(Site) - p3(Site))
\end{equation}
where \(p1, p2, p3\) are randomly selected parents from the mating pool, $F$ is the mutation factor, and $CR$ is the crossover probability. It emphasizes exploration by leveraging stochastic parent selection.

\paragraph{ $MSnum = 2$ (DE/best/1)} This strategy incorporates global guidance by using the best-known feasible solution ($Gbest$) in the population. When \(Site(d) = 1\), the offspring is generated as 
\begin{equation}
  O(Site) = fG(Site) + F \times (p2(Site) - p3(Site))
\end{equation}
where $fG$ is a matrix repeating decision variable of \(Gbest\). It prioritizes exploitation by directing perturbations toward the current optimal solution, with $fG$ falling back to feasible parents ($feasP$) if $Gbest$ is unavailable.

\paragraph{ $MSnum = 3$ (DE/cur-to-best/1)} This balanced strategy combines the current parent and the global best. For \(Site(d) = 1\), the update rule is:
\begin{equation}
\begin{split}
O(Site) &= p1(Site) + F \times (fG(Site) - p1(Site))\\
  &\quad + F \times (p2(Site) - p3(Site))  
  \end{split}
\end{equation}
where the term \(F \times (fG - p1)\) pulls the current parent toward $Gbest$, while the difference term \(F \times (p2 - p3)\) maintains exploration, striking a balance between exploitation and diversity.

\paragraph{$MSnum = 4$ (DE/rand-to-best/1)} Similar to DE/cur-to-best/1 but with an additional random permutation of parents. For \(Site(d) = 1\), it uses
\eref{eq_ms4} to generate offspring, where \(\widetilde{p1}\) is a randomly permuted version of p1. This permutation introduces extra stochasticity, enhancing robustness in noisy or multi-modal objective spaces.
\begin{equation}
\begin{split}
O(Site) &= \widetilde{p1}(Site) + F \times (fG(Site) - \widetilde{p1}(Site)) \\
&\quad + F \times (p2(Site) - p3(Site))
\end{split}
\label{eq_ms4}
\end{equation}

\subsubsection{Candidate constraint handling techniques}
CHTs determine how feasible and infeasible solutions are selected for next generation during \(envSelection(P,O,cht_g)\) in \alref{algDRLMCDE}, ensuring the population \(P'\) evolves toward both feasibility and optimality. Seven techniques are implemented as candidates.

\paragraph{ $CHTnum = 1$ (Death Penalty)} This technique penalizes infeasible solutions heavily to prioritize feasibility. For each solution, the fitness of each individual is defined as 
\begin{equation}
  fit = - (fs \times objs + \neg fs \times (CV + 1e10))
\end{equation}
where $fs$ is a binary indicator (1 if \(CV = \sum \max(0, cons) = 0\), else 0). The top $N$ solutions with the largest \(fit\) are selected, effectively eliminating highly infeasible solutions.

\paragraph{$CHTnum = 2$ (Weighted Penalty)} It uses a linear combination of objective values and CVs to balance feasibility and optimality. The fitness is calculated by  
\begin{equation}
  fit = - (objs +  c_p \times CV)
\end{equation}
where \(c_p = 0.3\) is a fixed penalty coefficient. 

\paragraph{$CHTnum = 3$ (Feasibility Rule 1)} This rule enforces a strict priority: feasible solutions are always preferred over infeasible ones. If there are at least $N$ feasible solutions, the top $N$ by objective value are selected. If not, all feasible solutions are retained, and the remaining slots are filled by infeasible solutions with the smallest violation. 

\paragraph{$CHTnum = 4$ (Feasibility Rule 2)} It follows a two-level ranking: first by feasibility (feasible solutions first) and then by objective value. The sorting key is ensuring feasible solutions are ordered by objective and infeasible ones are ordered by objective.

\paragraph{$CHTnum = 5$ (Tournament CHT)} This stochastic technique uses pairwise comparisons to select solutions. For each of $N$ selection steps, two solutions are randomly chosen; the winner is: (1) the feasible one if only one is feasible; (2) the one with better objective if both are feasible; (3) the one with smaller violation if both are infeasible. It introduces noise to avoid local optima, promoting diversity.

\paragraph{$CHTnum = 6$ (Epsilon Feasibility Ranking)} It relaxes the feasibility definition using a threshold \(\epsilon = 0.05\), treating solutions with \(violation \leq \epsilon\) as "epsilon-feasible". The sorting key allows solutions slightly violating constraints to compete with strictly feasible ones, useful for problems with tight constraints.

\paragraph{$CHTnum = 7$ (Stochastic Ranking)} This probabilistic technique balances objective and violation-based ranking. With probability \(Pf = 0.45\), solutions are compared by objective value; with probability \(1-Pf\), they are compared by violation.

\section{Experimental studies and analysis}
\label{sec_exp}
This section develops multiple experiments to demonstrate the effectiveness of SuperDE. Experiments are studied and analyzed in \sref{secE1}, \sref{secE2} and \sref{secE3} separately for the following research questions (RQs):
\begin{itemize}
  \item RQ1: How does the performance of SuperDE compare with advanced DE variants? See \sref{secE1}.
  \item RQ2: How well does SuperDE generalize to unseen problem instances? See \sref{secE2}. 
  \item RQ3: Has SuperDE learned a good policy, and have the two automatically configured components played a role where 1+1$>$2? See \sref{secE3}.
\end{itemize}

\subsection{Benchmark problems and performance metrics}
\label{secProblems}
Over the past two decades, numerous benchmark problem suites have been proposed to evaluate EAs. One of the most renowned is the CEC2010 benchmark \cite{mallipeddi2010problem}. This suite comprises 18 scalable functions, where the detailed description is presented in \tref{tabCEC2010}. CEC2010 serves as the training problem suite for SuperDE, while it is also used, with different optima, as the testing problem suite for RQ1, denoted as C2010\_F and C2010E\_F, respectively. For RQ2, 28 benchmark constrained optimization problems from the CEC2017 competition are utilized, denoted as C2017\_F1 to C2017\_F28, covering a wide range of constraints \cite{wu2017problem}. Additionally, a simpler benchmark set from G2000, consisting of thirteen functions labeled g01 to g13, is also used \cite{runarsson2000stochastic}. Furthermore, the BBOB2022 test suite, proposed at the Genetic and Evolutionary Computation Conference 2022, includes 54 non-linearly constrained test functions and is also employed for RQ2 \cite{dufosse2022building}. For RQ3, both CEC2017 and G2000 are utilized.

\tref{tabCEC2010} provides the detailed description of CEC2010 problems. 
\begin{table*}[h] \tiny
\centering
\caption{Details of 18 problems in CEC2010. $E$ and $I$ are the number of equality and inequality constraints separately, $D$ is the number of decision variables, $fsr$ is the estimated ratio between the feasible region and the search space.}
\begin{tabular}{ccccccc}
\toprule
\multirow{2}{*}{\makecell{Problem}} &\multirow{2}{*}{\makecell{Search Range}} & \multirow{2}{*}{\makecell{Type of\\Objective}} & \multicolumn{2}{c}{\makecell{Number of Constraints}} & \multicolumn{2}{c}{\makecell{$fsr$}} \\
\cmidrule(lr){4-5} \cmidrule(lr){6-7} 
& & & $E$ & $I$ & 10D & 30D  \\
\midrule
C2010\_F01 & $[0,10]^D$ & Non Separable & 0 & \makecell{2 \\ Non Separable} & 0.997689 & 1.000000 \\
\midrule
C2010\_F02 & $[-5.12,5.12]^D$ & Separable & \makecell{1 \\ Separable} & \makecell{2 \\ Separable} & 0.000000 & 0.000000 \\
\midrule
C2010\_F03 & $[-1000,1000]^D$ & Non Separable & \makecell{1 \\ Non Separable} & 0 & 0.000000 & 0.000000 \\
\midrule
C2010\_F04 & $[-50,50]^D$ & Separable & \makecell{4 \\ 2 Non Separable, 2 \\  Separable} & 0 & 0.000000 & 0.000000 \\
\midrule
C2010\_F05 & $[-600,600]^D$ & Separable & \makecell{2 \\ Separable} & 0 & 0.000000 & 0.000000 \\
\midrule
C2010\_F06 & $[-600,600]^D$ & Separable & \makecell{2 \\ Rotated} & 0 & 0.000000 & 0.000000 \\
\midrule
C2010\_F07 & $[-140,140]^D$ & Non Separable & 0 & \makecell{1 \\ Separable} & 0.505123 & 0.503725 \\
\midrule
C2010\_F08 & $[-140,140]^D$ & Non Separable & 0 & \makecell{1 \\ Rotated} & 0.379512 & 0.375278 \\
\midrule
C2010\_F09 & $[-500,500]^D$ & Non Separable & \makecell{1 \\ Separable} & 0 & 0.000000 & 0.000000 \\
\midrule
C2010\_F10 & $[-500,500]^D$ & Non Separable & \makecell{1 \\ Rotated} & 0 & 0.000000 & 0.000000 \\
\midrule
C2010\_F11 & $[-100,100]^D$ & Rotated & \makecell{1 \\ Non Separable} & 0 & 0.000000 & 0.000000 \\
\midrule
C2010\_F12 & $[-1000,1000]^D$ & Separable & \makecell{1 \\ Non Separable} & \makecell{1 \\ Separable} & 0.000000 & 0.000000 \\
\midrule
C2010\_F13 & $[-500,500]^D$ & Separable & 0 & \makecell{3 \\ 2 Separable, 1 Non \\ Separable} & 0.000000 & 0.000000 \\
\midrule
C2010\_F14 & $[-1000,1000]^D$ & Non Separable & 0 & \makecell{3 \\ Separable} & 0.003112 & 0.006123 \\
\midrule
C2010\_F15 & $[-1000,1000]^D$ & Non Separable & 0 & \makecell{3 \\ Rotated} & 0.003210 & 0.006023 \\
\midrule
C2010\_F16 & $[-10,10]^D$ & Non Separable & \makecell{2 \\ Separable} & \makecell{2 \\ 1 Separable, 1 Non \\ Separable} & 0.000000 & 0.000000 \\
\midrule
C2010\_F17 & $[-10,10]^D$ & Non Separable & \makecell{1 \\ Separable} & \makecell{2 \\ Non Separable} & 0.000000 & 0.000000 \\
\midrule
C2010\_F18 & $[-50,50]^D$ & Non Separable & \makecell{1 \\ Separable} & \makecell{1 \\ Separable} & 0.000010 & 0.000000 \\
\bottomrule
\end{tabular}
\label{tabCEC2010}
\end{table*}

Performance metrics for COPs assess algorithm performance based on feasibility and solution quality. For problems with feasible solutions, the algorithm with the smallest average objective value across 31 independent runs is deemed superior. For problems without feasible solutions, the algorithm with the smallest average CV across all solutions in each run is considered better. This dual metric framework ensures a fair and targeted assessment, aligning with the core goals of constrained optimization: achieving feasibility first, then enhancing solution quality.

\subsection{Competitor algorithms}
\label{secCompetitors}
For RQ1 and RQ2, DE \cite{storn1997differential} with death penalty as CHT is selected as the first competitor because the DRL is applied to dynamically configure DE in our proposed framework. Second, several advance DE variants for COPs, including CMODE \cite{wang2012combining}, CAMDE \cite{xu2019adaptive}, IepsilonJADE \cite{yi2023enhanced}, and C2ODE \cite{wang2018composite}, are implemented as competitive algorithms. In addition, CW \cite{cai2006multiobjective}, as a classical DE algorithm for COPs, is selected as competitors as well. 

For RQ3, we developed an ablation study for SuperDE. The competitors include 3 kinds of SuperDE variants, denoted as SuperDE1-3. In SuperDE1, the Agent configures the CHT only during evolution, while the MS is set to rand/1 ($MSnum = 1$). In SuperDE2, the Agent configures the MS only during evolution, while the CHT is set as death penalty ($CHTnum=1$). SuperDE3 is instantiated as a random configuration baseline, where the agent in SuperDE is replaced with randomly selecting the two components. 

\subsection{Results and Analysis for RQ1}
\label{secE1}
The optimization results on 10D CEC2010 extension are presented in \tref{tabRQ1C10E_minobj} (average minimum objective values) and \tref{tabRQ1C10E_mincv} (average minimum CVs on problems where no algorithm found feasible solutions). \tref{tabRQ1C10E_minobj} reveals that SuperDE exhibits strong performance in finding feasible solutions and achieving lower objective values compared to advanced DE variants. Specifically, in C2010E\_F7, SuperDE achieved an objective value of 4.5982e+2, which is significantly lower than all competitors, highlighting its ability to navigate constrained search spaces effectively. Similarly, C2010E\_F16 stands out as a problem where only SuperDE found a feasible solution (5.7471e-1), while all other algorithms returned NaN, underscoring its superior feasibility rate.

In problems where multiple algorithms found feasible solutions, SuperDE consistently delivered competitive or better results. For example, in C2010E\_F13, SuperDE's objective value was the lowest among all tested algorithms, outperforming DE, CMODE and CAMDE. In C2010E\_F18, SuperDE was significantly lower than CMODE and CAMDE and comparable to C2ODE.

Statistical significance, as indicated by the Wilcoxon signed-rank test further supports SuperDE's effectiveness. Most competitors showed more "-" results than "+" results. For example, DE and IepsilonJADE had 3 and 5 "-" results, respectively, with no "+" results, while CAMDE and CMODE - two of the most competitive variants - still had 6 and 5 "-" results, respectively, confirming SuperDE's statistical superiority in most cases.

For problems where no algorithm found feasible solutions, shown in \tref{tabRQ1C10E_mincv}, SuperDE demonstrated exceptional ability to minimize CVs, a critical metric for constrained optimization. In C2010E\_F5, SuperDE's CV was the lowest among all algorithms, outperforming DE, CW and CAMDE. Similarly, in C2010E\_F11, SuperDE achieved a CV of 1.2498e+2, which was significantly lower than DE, CW and IepsilonJADE, highlighting its robustness in reducing CVs even when feasibility is not attained. Notably, in C2010E\_F8 and C2010E\_F10, SuperDE's CV values are comparable to the CAMDE and CMODE, with no significant statistical difference ("="), further validating its ability to balance exploration and exploitation in challenging infeasible regions. The Wilcoxon test results for these problems reinforce that SuperDE had fewer "-" results and more "=" results compared to most variants, with only CMODE and CAMDE showing marginal competitiveness in specific cases.

\begin{figure*}
  \centering
  \subfloat[]{\includegraphics[width=0.25\textwidth]{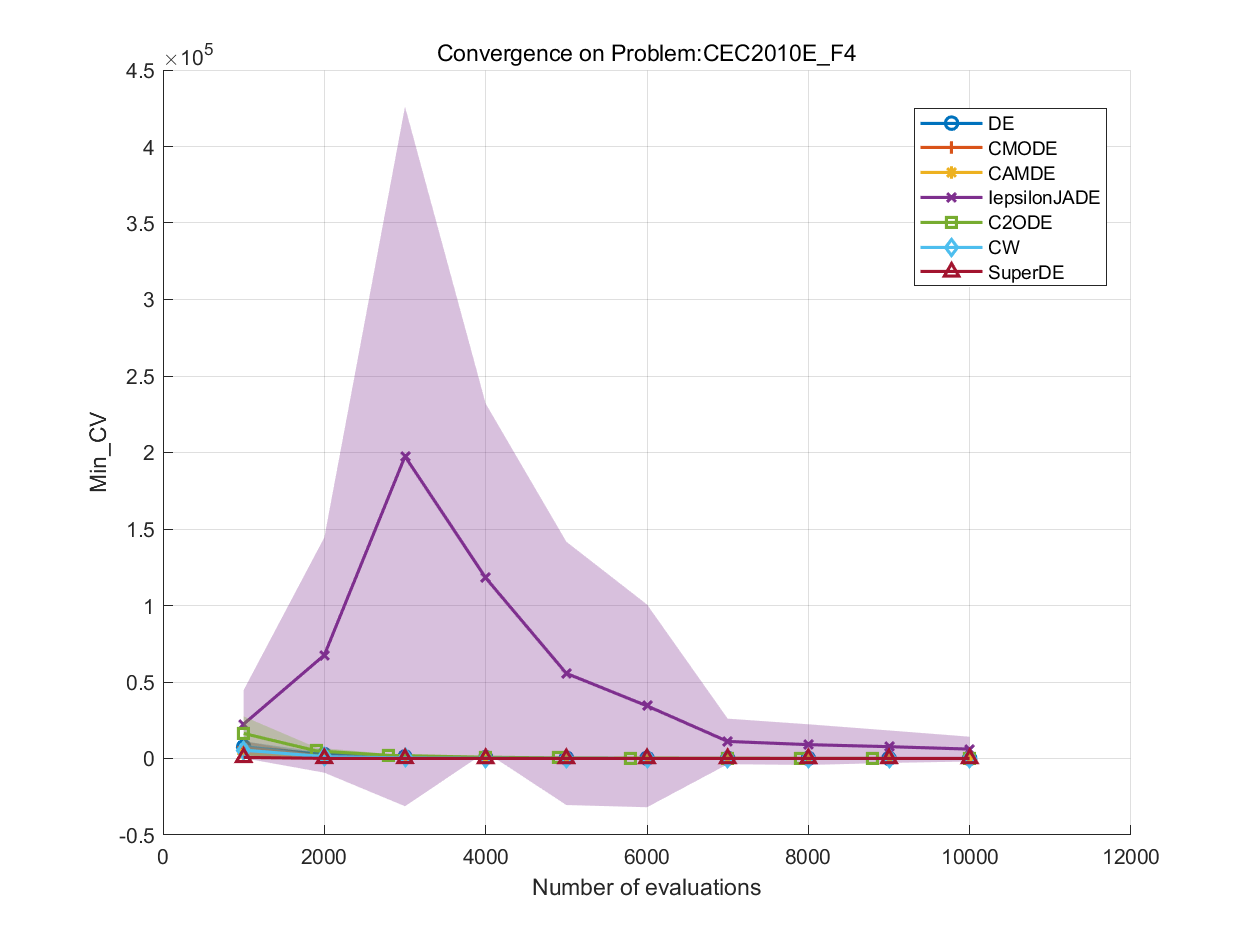}} 
  \subfloat[]{\includegraphics[width=0.25\textwidth]{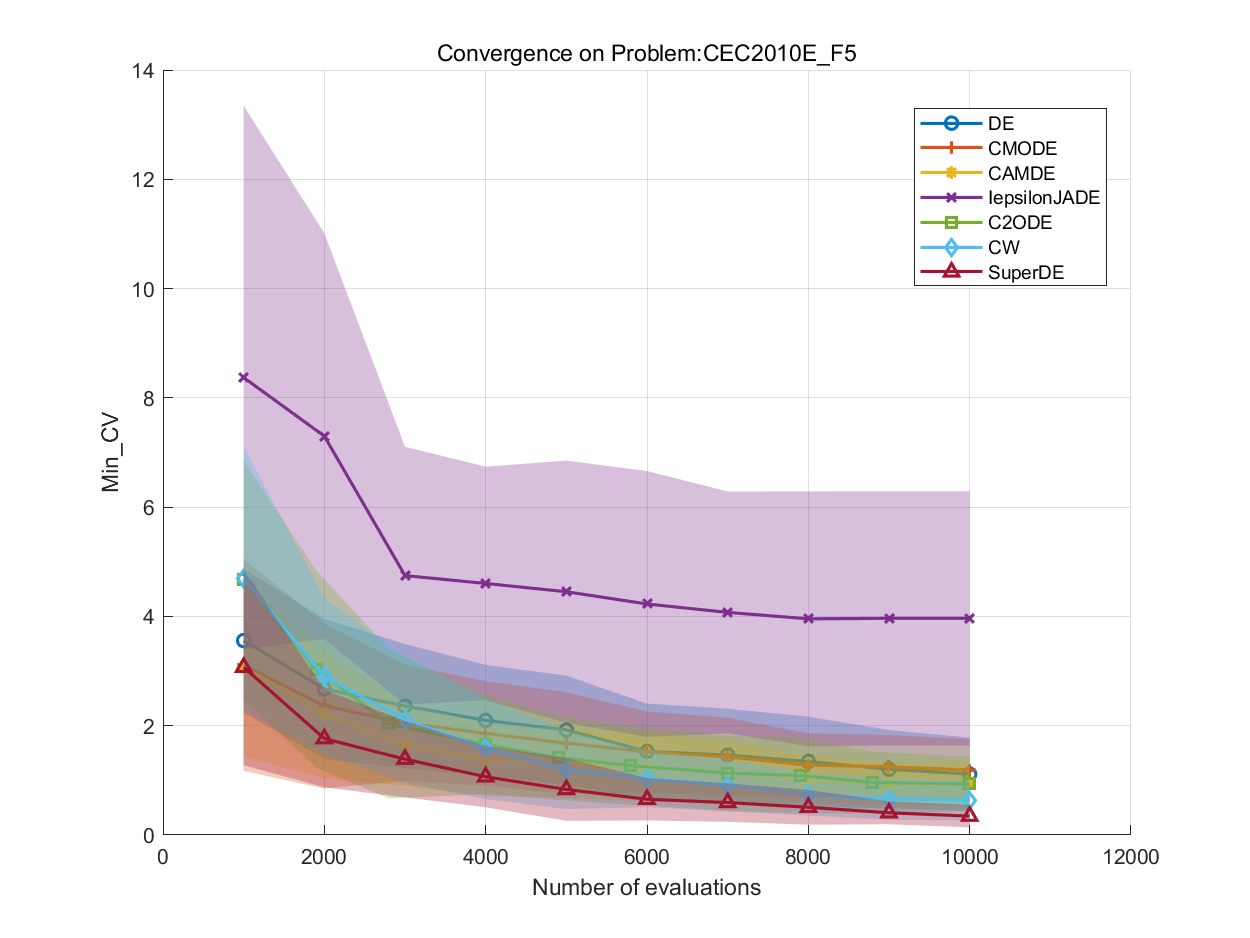}} 
  \subfloat[]{\includegraphics[width=0.25\textwidth]{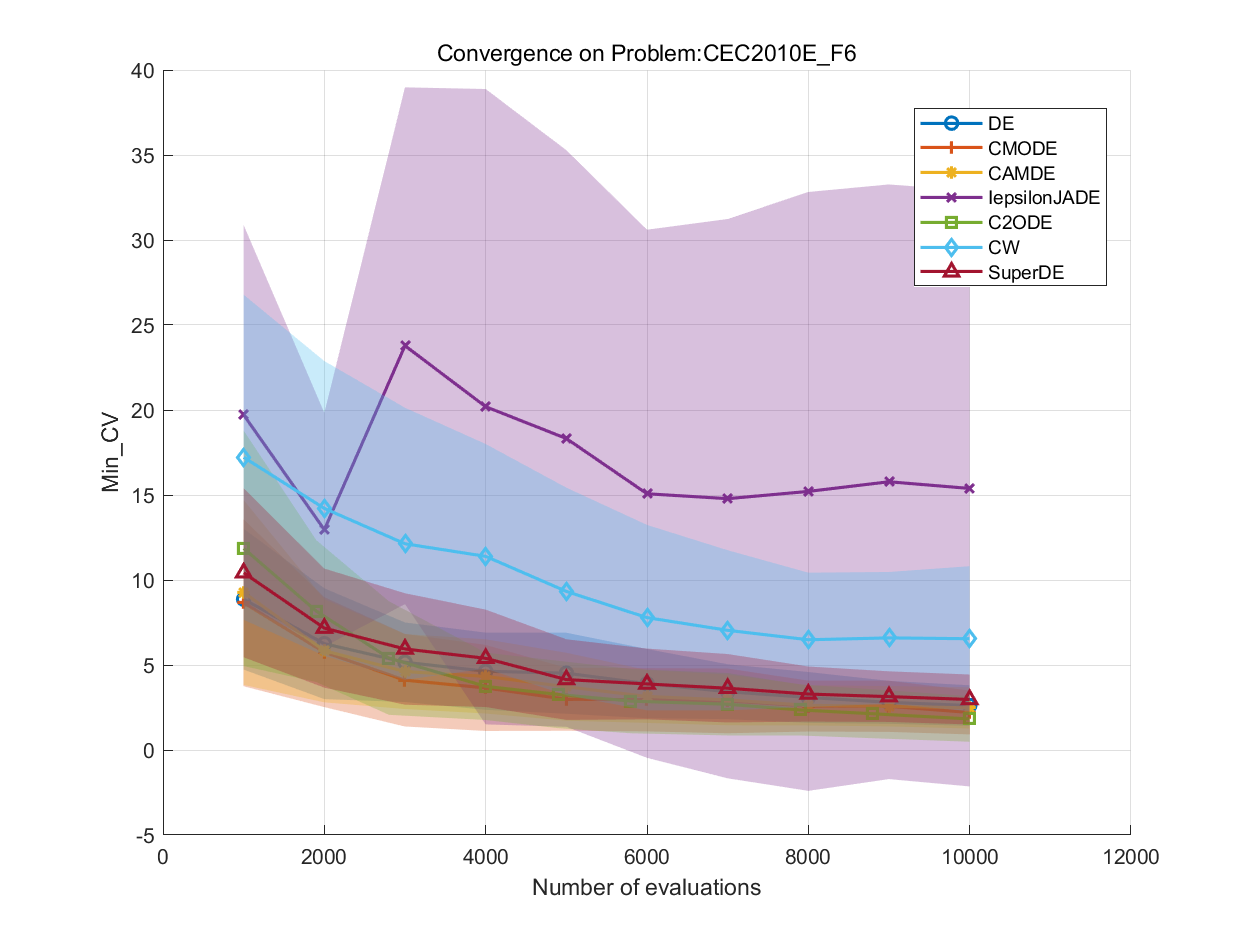}} \\
  \subfloat[]{\includegraphics[width=0.25\textwidth]{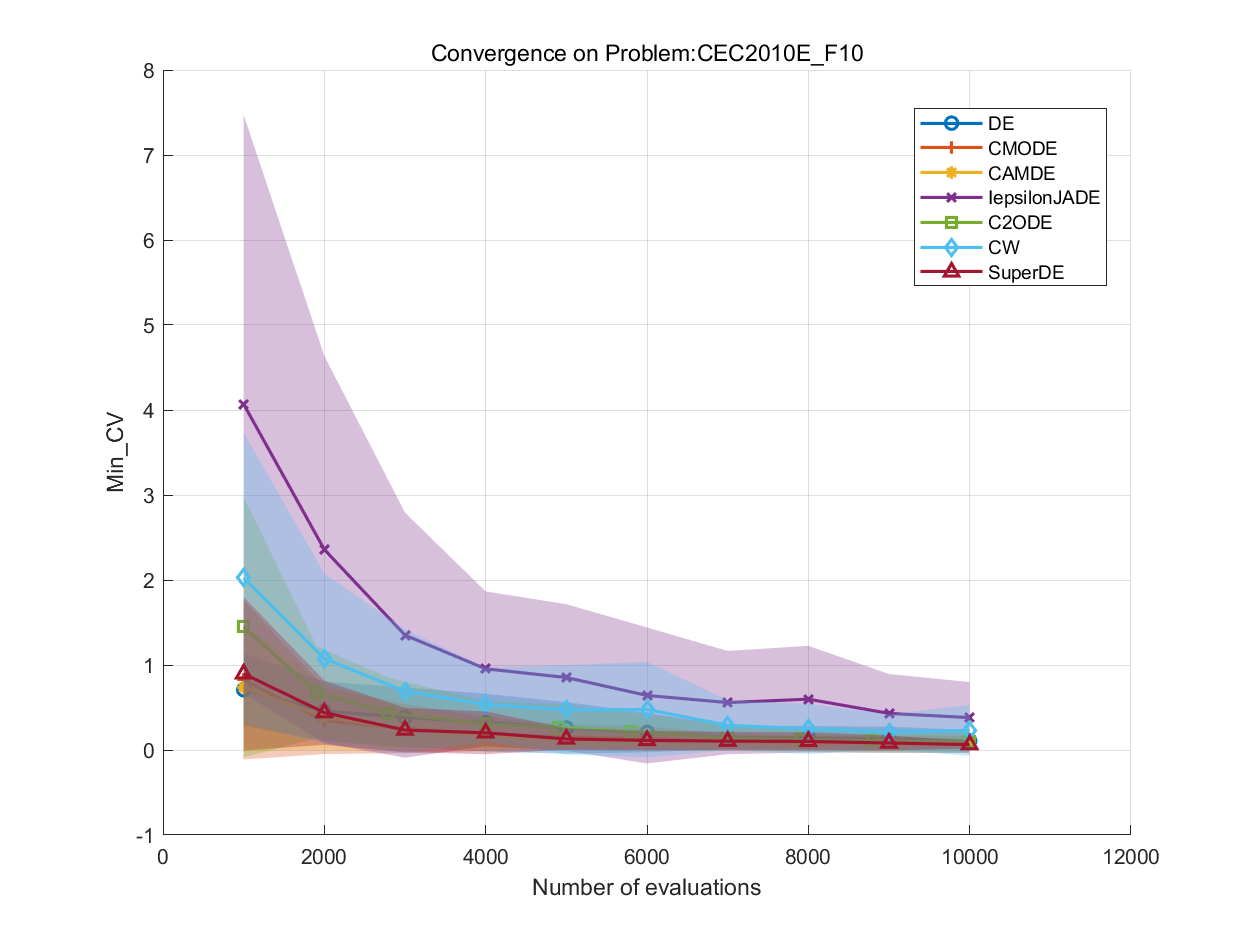}} 
  \subfloat[]{\includegraphics[width=0.25\textwidth]{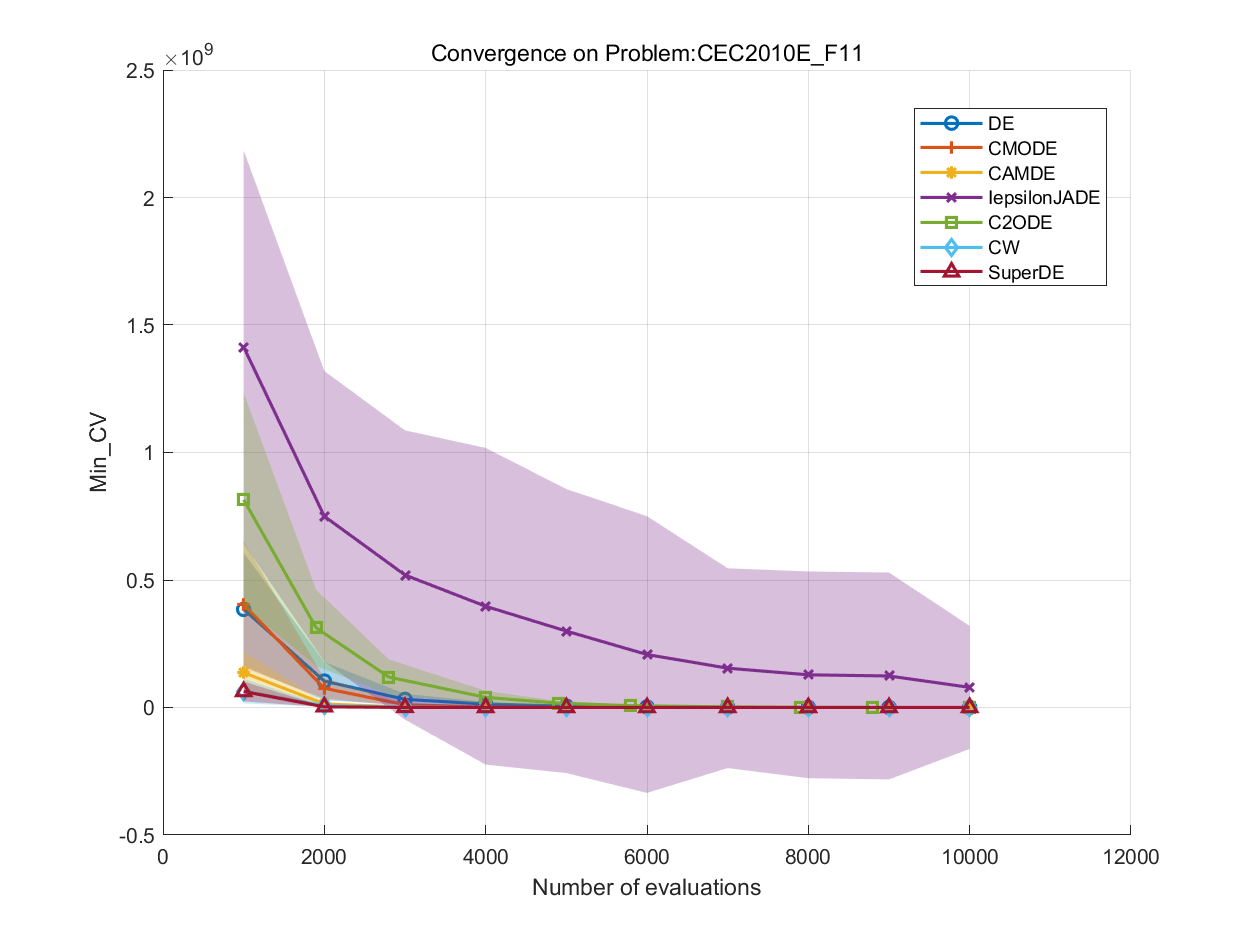}} 
  \subfloat[]{\includegraphics[width=0.25\textwidth]{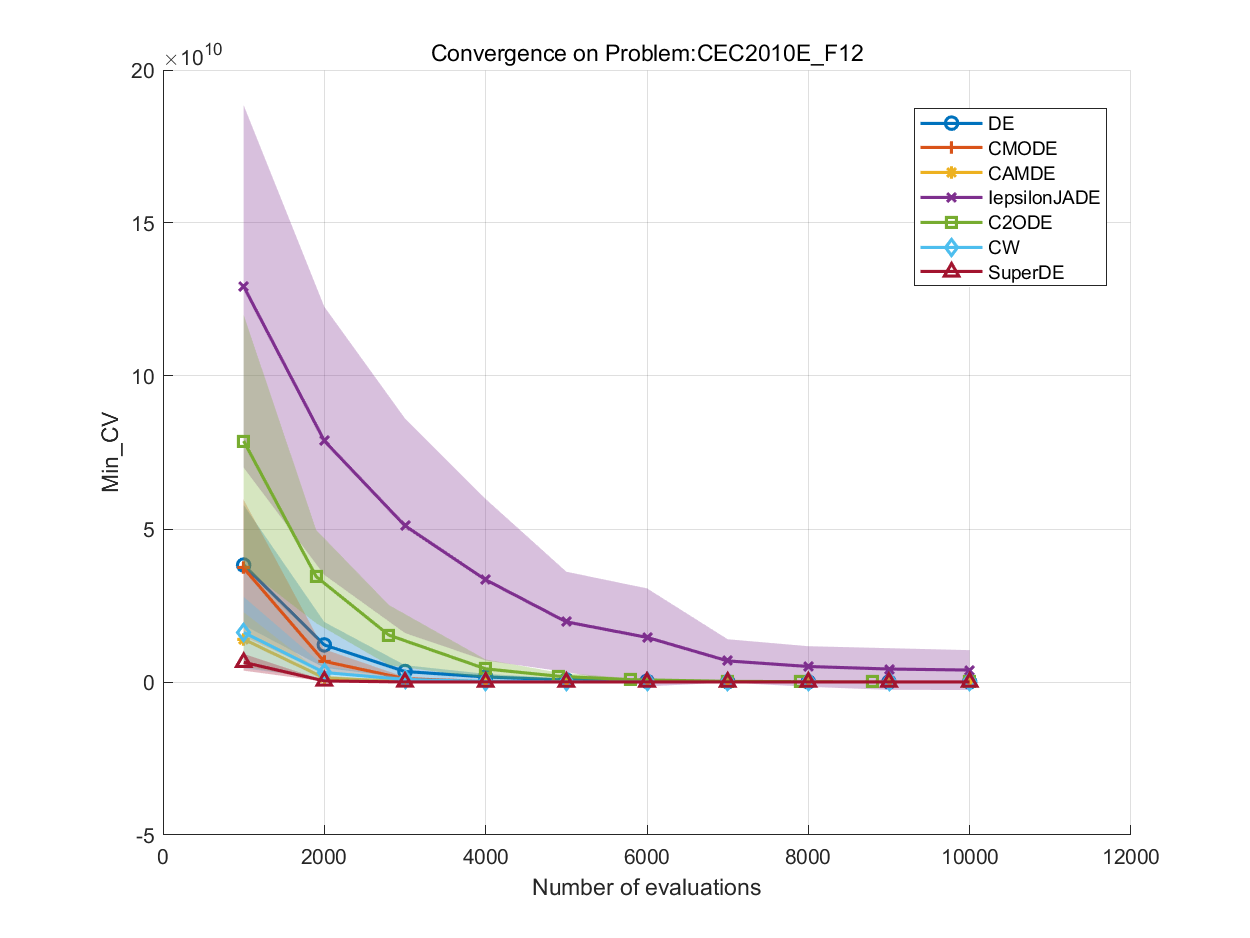}} \\
  \caption{CV convergence of different algorithms on CEC2010 extension F4, F5, F6, F10, F11, and F12.}
  \label{figCVTrend}
\end{figure*}

For further analyze the convergence of different algorithms, the CV results trend on CEC2010 extension F4, F5, F6, F10, F11, and F12 are selected to be presented in \fref{figCVTrend}. It can be concluded that SuperDE, depiceted by dark red line, shows better convergence to minimum CV, indicating that it is more likely to find optimum feasible solution. The trend of IepsilonJADE are unsatisfactory, while the other algorithms all have relatively stable convergence trend, with SuperDE showing the most outstanding performance.

Overall, the results for RQ1 demonstrate that SuperDE outperforms these competitors in key aspects of constrained optimization. It achieves lower objective values in feasible regions, and minimizes CVs in infeasible cases. Statistical tests confirm that these advantages are significant across most problems, establishing SuperDE as a competitive method to state-of-the-art DE variants for the CEC2010 benchmark suite.

\begin{table*}[htbp] \tiny
  \centering
  \caption{10D CEC2010 extension experimental results (average minimum objective values and standard deviations) for RQ1. "NaN" means no feasible solution, while "-" means worse than SuperDE, "+" means better, and "=" means no significant difference. The best results are highlighted with gray background.  All algorithm haven't found feasible solution on C2010E\_F4-6, F8 and F10-12, whose results are not displayed here.}
  \begin{tabular}{cccccccc}
    \toprule
    \textbf{Problem} & \textbf{DE} & \textbf{CW} & \textbf{CMODE} & \textbf{CAMDE} & \textbf{IepsilonJADE} & \textbf{C2ODE} & \textbf{SuperDE} \\
    \midrule
    C2010E\_F1 & \multicolumn{1}{p{6.1em}}{-4.0342e-1 \newline{}(3.01e-2) -} & \multicolumn{1}{p{6.1em}}{-2.7145e-1 \newline{}(2.79e-2) -} & \multicolumn{1}{p{6.1em}}{-5.3049e-1 \newline{}(3.15e-2) =} & \multicolumn{1}{p{6.1em}}{\cellcolor[rgb]{ .906,  .902,  .902}-5.9636e-1 \newline{}(3.88e-2) +} & \multicolumn{1}{p{6.1em}}{-3.3396e-1 \newline{}(9.35e-2) -} & \multicolumn{1}{p{6.1em}}{-5.1115e-1 \newline{}(2.85e-2) =} & \multicolumn{1}{p{6.1em}}{-5.1430e-1 \newline{}(7.67e-2)} \\
    C2010E\_F2 & \multicolumn{1}{p{6.1em}}{NaN \newline{}(NaN)} & \multicolumn{1}{p{6.1em}}{NaN \newline{}(NaN)} & \multicolumn{1}{p{6.1em}}{\cellcolor[rgb]{ .906,  .902,  .902}2.8613e+0 \newline{}(1.71e+0) =} & \multicolumn{1}{p{6.1em}}{\cellcolor[rgb]{ .906,  .902,  .902}3.2430e+0 \newline{}(1.28e+0) =} & \multicolumn{1}{p{6.1em}}{NaN \newline{}(NaN)} & \multicolumn{1}{p{6.1em}}{\cellcolor[rgb]{ .906,  .902,  .902}2.0213e+0 \newline{}(3.26e-1) =} & \multicolumn{1}{p{6.1em}}{\cellcolor[rgb]{ .906,  .902,  .902}1.5376e+0 \newline{}(9.71e-1)} \\
    C2010E\_F3 & \multicolumn{1}{p{6.1em}}{\cellcolor[rgb]{ .906,  .902,  .902}4.5287e-3 \newline{}(6.95e-3) =} & \multicolumn{1}{p{6.1em}}{NaN \newline{}(NaN)} & \multicolumn{1}{p{6.1em}}{NaN \newline{}(NaN)} & \multicolumn{1}{p{6.1em}}{8.9814e+6 \newline{}(3.17e+4) -} & \multicolumn{1}{p{6.1em}}{8.9640e+6 \newline{}(0.00e+0) -} & \multicolumn{1}{p{6.1em}}{\cellcolor[rgb]{ .906,  .902,  .902}3.1880e+6 \newline{}(4.37e+6) =} & \multicolumn{1}{p{6.1em}}{\cellcolor[rgb]{ .906,  .902,  .902}8.4942e-1 \newline{}(1.45e+0)} \\
    C2010E\_F7 & \multicolumn{1}{p{6.1em}}{1.1689e+9 \newline{}(8.07e+8) -} & \multicolumn{1}{p{6.1em}}{1.2529e+7 \newline{}(1.42e+7) -} & \multicolumn{1}{p{6.1em}}{3.8073e+7 \newline{}(2.07e+7) -} & \multicolumn{1}{p{6.1em}}{1.5731e+4 \newline{}(1.30e+4) -} & \multicolumn{1}{p{6.1em}}{1.0107e+9 \newline{}(2.51e+9) -} & \multicolumn{1}{p{6.1em}}{2.1455e+6 \newline{}(1.46e+6) -} & \multicolumn{1}{p{6.1em}}{\cellcolor[rgb]{ .906,  .902,  .902}4.5982e+2 \newline{}(1.17e+3)} \\
    C2010E\_F9 & \multicolumn{1}{p{6.1em}}{NaN \newline{}(NaN)} & \multicolumn{1}{p{6.1em}}{NaN \newline{}(NaN)} & \multicolumn{1}{p{6.1em}}{\cellcolor[rgb]{ .906,  .902,  .902}1.7431e+13 \newline{}(0.00e+0)} & \multicolumn{1}{p{6.1em}}{NaN \newline{}(NaN)} & \multicolumn{1}{p{6.1em}}{5.5856e+13 \newline{}(1.39e+11)} & \multicolumn{1}{p{6.1em}}{5.6238e+13 \newline{}(1.29e+11)} & \multicolumn{1}{p{6.1em}}{NaN \newline{}(NaN)} \\
    C2010E\_F13 & \multicolumn{1}{p{6.1em}}{-4.2342e+1 \newline{}(3.17e+0) -} & \multicolumn{1}{p{6.1em}}{-3.1230e+1 \newline{}(3.42e+0) -} & \multicolumn{1}{p{6.1em}}{-4.4562e+1 \newline{}(3.20e+0) -} & \multicolumn{1}{p{6.1em}}{-1.6481e+1 \newline{}(1.34e+1) -} & \multicolumn{1}{p{6.1em}}{-4.5710e+1 \newline{}(8.29e+0) -} & \multicolumn{1}{p{6.1em}}{-4.5344e+1 \newline{}(3.08e+0) -} & \multicolumn{1}{p{6.1em}}{\cellcolor[rgb]{ .906,  .902,  .902}-6.0443e+1 \newline{}(3.46e+0)} \\
    C2010E\_F14 & \multicolumn{1}{p{6.1em}}{NaN \newline{}(NaN)} & \multicolumn{1}{p{6.1em}}{7.5720e+11 \newline{}(9.73e+11) -} & \multicolumn{1}{p{6.1em}}{1.7985e+13 \newline{}(1.06e+13) -} & \multicolumn{1}{p{6.1em}}{1.8279e+13 \newline{}(9.03e+12) -} & \multicolumn{1}{p{6.1em}}{7.8718e+13 \newline{}(1.32e+14) -} & \multicolumn{1}{p{6.1em}}{3.7147e+13 \newline{}(2.50e+13) -} & \multicolumn{1}{p{6.1em}}{\cellcolor[rgb]{ .906,  .902,  .902}1.1009e+10 \newline{}(2.18e+10)} \\
    C2010E\_F15 & \multicolumn{1}{p{6.1em}}{NaN \newline{}(NaN)} & \multicolumn{1}{p{6.1em}}{\cellcolor[rgb]{ .906,  .902,  .902}4.1771e+12 \newline{}(3.20e+12) +} & \multicolumn{1}{p{6.1em}}{1.3129e+14 \newline{}(6.50e+13) -} & \multicolumn{1}{p{6.1em}}{8.0398e+13 \newline{}(4.63e+13) -} & \multicolumn{1}{p{6.1em}}{1.6588e+14 \newline{}(2.64e+14) =} & \multicolumn{1}{p{6.1em}}{3.1086e+14 \newline{}(1.87e+14) -} & \multicolumn{1}{p{6.1em}}{1.1101e+13 \newline{}(4.76e+13)} \\
    C2010E\_F16 & \multicolumn{1}{p{6.1em}}{NaN \newline{}(NaN)} & \multicolumn{1}{p{6.1em}}{NaN \newline{}(NaN)} & \multicolumn{1}{p{6.1em}}{NaN \newline{}(NaN)} & \multicolumn{1}{p{6.1em}}{NaN \newline{}(NaN)} & \multicolumn{1}{p{6.1em}}{NaN \newline{}(NaN)} & \multicolumn{1}{p{6.1em}}{NaN \newline{}(NaN)} & \multicolumn{1}{p{6.1em}}{\cellcolor[rgb]{ .906,  .902,  .902}5.7471e-1 \newline{}(2.50e-1)} \\
    C2010E\_F17 & \multicolumn{1}{p{6.1em}}{\cellcolor[rgb]{ .906,  .902,  .902}4.6103e+2 \newline{}(0.00e+0) =} & \multicolumn{1}{p{6.1em}}{NaN \newline{}(NaN)} & \multicolumn{1}{p{6.1em}}{\cellcolor[rgb]{ .906,  .902,  .902}5.3380e+2 \newline{}(0.00e+0) =} & \multicolumn{1}{p{6.1em}}{\cellcolor[rgb]{ .906,  .902,  .902}7.6166e+2 \newline{}(0.00e+0) =} & \multicolumn{1}{p{6.1em}}{\cellcolor[rgb]{ .906,  .902,  .902}3.5316e+1 \newline{}(0.00e+0) =} & \multicolumn{1}{p{6.1em}}{NaN \newline{}(NaN)} & \multicolumn{1}{p{6.1em}}{\cellcolor[rgb]{ .906,  .902,  .902}6.1167e+1 \newline{}(3.23e+1)} \\
    C2010E\_F18 & \multicolumn{1}{p{6.1em}}{\cellcolor[rgb]{ .906,  .902,  .902}1.3378e+4 \newline{}(0.00e+0) =} & \multicolumn{1}{p{6.1em}}{NaN \newline{}(NaN)} & \multicolumn{1}{p{6.1em}}{1.2553e+4 \newline{}(4.95e+3) -} & \multicolumn{1}{p{6.1em}}{1.4865e+4 \newline{}(5.23e+3) -} & \multicolumn{1}{p{6.1em}}{NaN \newline{}(NaN)} & \multicolumn{1}{p{6.1em}}{8.5166e+3 \newline{}(0.00e+0) =} & \multicolumn{1}{p{6.1em}}{\cellcolor[rgb]{ .906,  .902,  .902}1.6360e+3 \newline{}(7.48e+2)} \\
    \midrule
    \multicolumn{1}{c}{+/-/=} & 0/3/3 & 1/4/0 & 0/5/3 & 1/6/2 & 0/5/2 & 0/4/4 &  \\
    \bottomrule
    \end{tabular}%
  \label{tabRQ1C10E_minobj}%
\end{table*}%

\begin{table*}[htbp] \tiny
  \centering
  \caption{10D CEC2010 extension experimental results (average minimum CV and standard deviations) for RQ1 on problems where no algorithm found feasible solution}
    \begin{tabular}{cccccccc}
    \toprule
    \textbf{Problem} & \textbf{DE} & \textbf{CW} & \textbf{CMODE} & \textbf{CAMDE} & \textbf{IepsilonJADE} & \textbf{C2ODE} & \textbf{SuperDE} \\
    \midrule
    C2010E\_F4 & \multicolumn{1}{p{6.1em}}{4.1664e+1 \newline{}(1.67e+1) -} & \multicolumn{1}{p{6.1em}}{4.3189e+1 \newline{}(3.93e+1) -} & \multicolumn{1}{p{6.1em}}{3.5785e+0 \newline{}(1.80e+0) =} & \multicolumn{1}{p{6.1em}}{\cellcolor[rgb]{ .906,  .902,  .902}1.2259e+0 \newline{}(6.87e-1) =} & \multicolumn{1}{p{6.1em}}{6.1397e+3 \newline{}(8.10e+3) -} & \multicolumn{1}{p{6.1em}}{5.7196e+1 \newline{}(1.59e+1) -} & \multicolumn{1}{p{6.1em}}{\cellcolor[rgb]{ .906,  .902,  .902}6.4804e+0 \newline{}(8.44e+0)} \\
    C2010E\_F5 & \multicolumn{1}{p{6.1em}}{1.1157e+0 \newline{}(6.68e-1) -} & \multicolumn{1}{p{6.1em}}{6.3894e-1 \newline{}(3.70e-1) -} & \multicolumn{1}{p{6.1em}}{1.1879e+0 \newline{}(5.68e-1) -} & \multicolumn{1}{p{6.1em}}{9.3698e-1 \newline{}(3.95e-1) -} & \multicolumn{1}{p{6.1em}}{3.9655e+0 \newline{}(2.33e+0) -} & \multicolumn{1}{p{6.1em}}{9.3400e-1 \newline{}(4.98e-1) -} & \multicolumn{1}{p{6.1em}}{\cellcolor[rgb]{ .906,  .902,  .902}3.4597e-1 \newline{}(2.07e-1)} \\
    C2010E\_F6 & \multicolumn{1}{p{6.1em}}{2.6145e+0 \newline{}(1.19e+0) =} & \multicolumn{1}{p{6.1em}}{6.5456e+0 \newline{}(4.27e+0) -} & \multicolumn{1}{p{6.1em}}{2.2078e+0 \newline{}(1.29e+0) +} & \multicolumn{1}{p{6.1em}}{2.4759e+0 \newline{}(1.21e+0) =} & \multicolumn{1}{p{6.1em}}{1.5384e+1 \newline{}(1.75e+1) -} & \multicolumn{1}{p{6.1em}}{\cellcolor[rgb]{ .906,  .902,  .902}1.8267e+0 \newline{}(1.34e+0) +} & \multicolumn{1}{p{6.1em}}{2.9718e+0 \newline{}(1.47e+0)} \\
    C2010E\_F8 & \multicolumn{1}{p{6.1em}}{1.9034e-1 \newline{}(1.67e-1) -} & \multicolumn{1}{p{6.1em}}{4.8283e-1 \newline{}(3.72e-1) -} & \multicolumn{1}{p{6.1em}}{\cellcolor[rgb]{ .906,  .902,  .902}1.2650e-1 \newline{}(1.03e-1) =} & \multicolumn{1}{p{6.1em}}{\cellcolor[rgb]{ .906,  .902,  .902}1.1240e-1 \newline{}(1.33e-1) =} & \multicolumn{1}{p{6.1em}}{5.6388e-1 \newline{}(5.22e-1) -} & \multicolumn{1}{p{6.1em}}{2.1506e-1 \newline{}(1.75e-1) -} & \multicolumn{1}{p{6.1em}}{\cellcolor[rgb]{ .906,  .902,  .902}9.0769e-2 \newline{}(6.55e-2)} \\
    C2010E\_F10 & \multicolumn{1}{p{6.1em}}{\cellcolor[rgb]{ .906,  .902,  .902}1.0774e-1 \newline{}(1.27e-1) =} & \multicolumn{1}{p{6.1em}}{2.3190e-1 \newline{}(3.00e-1) -} & \multicolumn{1}{p{6.1em}}{\cellcolor[rgb]{ .906,  .902,  .902}8.6900e-2 \newline{}(8.17e-2) =} & \multicolumn{1}{p{6.1em}}{\cellcolor[rgb]{ .906,  .902,  .902}7.9396e-2 \newline{}(6.36e-2) =} & \multicolumn{1}{p{6.1em}}{3.8083e-1 \newline{}(4.18e-1) -} & \multicolumn{1}{p{6.1em}}{\cellcolor[rgb]{ .906,  .902,  .902}9.6878e-2 \newline{}(1.31e-1) =} & \multicolumn{1}{p{6.1em}}{\cellcolor[rgb]{ .906,  .902,  .902}6.3092e-2 \newline{}(4.69e-2)} \\
    C2010E\_F11 & \multicolumn{1}{p{6.1em}}{1.3683e+5 \newline{}(7.88e+4) -} & \multicolumn{1}{p{6.1em}}{1.4835e+4 \newline{}(2.31e+4) -} & \multicolumn{1}{p{6.1em}}{1.4916e+3 \newline{}(8.87e+2) -} & \multicolumn{1}{p{6.1em}}{1.7407e+2 \newline{}(1.20e+2) -} & \multicolumn{1}{p{6.1em}}{7.8866e+7 \newline{}(2.41e+8) -} & \multicolumn{1}{p{6.1em}}{1.5848e+5 \newline{}(9.18e+4) -} & \multicolumn{1}{p{6.1em}}{\cellcolor[rgb]{ .906,  .902,  .902}1.2498e+2 \newline{}(1.57e+2)} \\
    C2010E\_F12 & \multicolumn{1}{p{6.1em}}{9.8867e+6 \newline{}(7.43e+6) -} & \multicolumn{1}{p{6.1em}}{1.5088e+7 \newline{}(1.77e+7) -} & \multicolumn{1}{p{6.1em}}{1.6676e+4 \newline{}(1.45e+4) +} & \multicolumn{1}{p{6.1em}}{\cellcolor[rgb]{ .906,  .902,  .902}5.0274e+2 \newline{}(6.56e+2) +} & \multicolumn{1}{p{6.1em}}{3.9005e+9 \newline{}(6.51e+9) -} & \multicolumn{1}{p{6.1em}}{1.3326e+7 \newline{}(9.08e+6) -} & \multicolumn{1}{p{6.1em}}{2.1639e+4 \newline{}(9.36e+4)} \\
    \midrule
    +/-/= & 0/5/2 & 0/7/0 & 2/2/3 & 1/2/4 & 0/7/0 & 1/5/1 &  \\
    \bottomrule
    \end{tabular}%
  \label{tabRQ1C10E_mincv}%
\end{table*}%

\subsection{Results and Analysis for RQ2}
\label{secE2}
To address RQ2 (How well does SuperDE generalize to unseen problem instances?), we evaluate the performance of SuperDE on CEC2017, G2000 and BBOB2022. Results on CEC2017 are presented in \tref{tabRQ2C17_1} (minimum objective values) and \tref{tabRQ2C17_2} (CV, for problems where SuperDE failed to find feasible solutions on CEC2017). \tref{tabRQ2G1} presents the obtained minimum objective value results on G2000. 

As shown in \tref{tabRQ2C17_1}, SuperDE demonstrates strong ability to find feasible solutions across most CEC2017 problems. The fewest problems (C2017\_F7, F11, F17-F19, F26-F28) resulted in NaN by SuperDE, indicating high feasibility rates compared to competitors. DE, CW, and IepsilonJADE returned NaN for over half the problems. Specifically, in C2017\_F6, all comparative algorithms returned NaN, while SuperDE achieved a low objective value, highlighting its unique capability to handle this problem. In C2017\_F1-F5, C2017\_F8-F10, and C2017\_F12-F16, SuperDE consistently obtained lower objective values than most competitors. Wilcoxon test results at the bottom of \tref{tabRQ2C17_1} show that SuperDE outperformed most algorithms. DE, CW, IepsilonJADE, and C2ODE had 4-8 "-" results with few or no "+" results. CMODE and CAMDE, while more competitive, still had 9 and 8 "-" results, respectively, indicating SuperDE's statistical superiority in most cases.
As for problems where SuperDE failed to find feasible solutions, shown in \tref{tabRQ2C17_2}, the CV values of SuperDE remained competitive. In C2017\_F7, SuperDE's CV was better than 5 out of 6 competitors, with only C2ODE showing a lower CV. In C2017\_F17 and C2017\_F26, SuperDE's CV was comparable to top performers (e.g., CMODE and CAMDE) and significantly lower than DE, CW, and IepsilonJADE, demonstrating its ability to minimize CVs even when feasibility is not achieved.

As shown in \tref{tabRQ2G1}, the experimental results on G2000 suite further validates SuperDE's better generalization to classical constrained problems. SuperDE have found feasible solutions for all 13 G2000 problems, while DE, CAMDE, and IepsilonJADE returns NaN for 1-3 problems. In g3, SuperDE's objective value is significantly lower than all competitors. In g4 and g8, SuperDE matches or approached the theoretical optimal values with minimal standard deviations, indicating high stability. In g1, g2, and g11, SuperDE performs comparably to top algorithms CMODE and C2ODE with "=" results, despite these competitors showing marginal "+" advantages in specific cases. The Wilcoxon test confirms that SuperDE outperforms DE, IepsilonJADE, C2ODE, and CW (0-1 "+" results against SuperDE). While CMODE and CAMDE have 1-2 "+" results, 7-8 "-" results highlightins the better performance of SuperDE.

The results on BBOB2022, presented in the appendix due to space constraints, further underscore the effectiveness and generalizability of our proposed SuperDE. In conclusion, SuperDE has demonstrated attractive generalizability to unseen problems (CEC2017 and G2000). On the one hand, SuperDE finds feasible solutions for more problems than competitors, including instances where all others failed.
On the other hand, SuperDE obtains consistently lower values in most problems, with small standard deviations indicating its stability. In addition, even when feasible solutions are not found, CV values obtained by SuperDE remains minimal compared to competitors. The Wilcoxon test confirms SuperDE outperforms or matches state-of-the-art algorithms in most cases. These results indicate that SuperDE effectively generalizes to diverse unseen constrained optimization problems.

\begin{table*}[htbp] \tiny
  \centering
  \caption{10D CEC2017 experimental results (average minimum objective values and standard deviations) for RQ2. "NaN" means no feasible solution, while "-" means worse than SuperDE, "+" means better, and "=" means no significant difference. All algorithm haven't found feasible solution on C2017\_F7, F11, F17-19 and F26-28, whose results are not displayed here.}
  \begin{tabular}{cccccccc}
    \toprule
    \textbf{Problem} & \textbf{DE} & \textbf{CW} & \textbf{CMODE} & \textbf{CAMDE} & \textbf{IepsilonJADE} & \textbf{C2ODE} & \textbf{SuperDE} \\
    \midrule
    C2017\_F1  & \multicolumn{1}{p{5.435em}}{2.5334e+2 \newline{}(9.25e+1) -} & \multicolumn{1}{p{5.435em}}{6.7059e+2 \newline{}(4.92e+2) -} & \multicolumn{1}{p{5.435em}}{5.1548e+2 \newline{}(1.78e+2) -} & \multicolumn{1}{p{5.435em}}{2.3884e+1 \newline{}(1.54e+1) -} & \multicolumn{1}{p{5.435em}}{9.5353e+2 \newline{}(2.58e+3) -} & \multicolumn{1}{p{5.435em}}{1.9789e+2 \newline{}(6.96e+1) -} & \multicolumn{1}{p{5.435em}}{\cellcolor[rgb]{ .906,  .902,  .902}9.2319e-1 \newline{}(2.89e+0)} \\
    C2017\_F2  & \multicolumn{1}{p{5.435em}}{3.3336e+2 \newline{}(9.28e+1) -} & \multicolumn{1}{p{5.435em}}{2.2096e+3 \newline{}(2.58e+3) -} & \multicolumn{1}{p{5.435em}}{3.2160e+2 \newline{}(1.04e+2) -} & \multicolumn{1}{p{5.435em}}{5.4320e+1 \newline{}(1.52e+1) -} & \multicolumn{1}{p{5.435em}}{2.4441e+3 \newline{}(4.91e+3) -} & \multicolumn{1}{p{5.435em}}{2.2978e+2 \newline{}(6.87e+1) -} & \multicolumn{1}{p{5.435em}}{\cellcolor[rgb]{ .906,  .902,  .902}7.3498e-1 \newline{}(3.92e+0)} \\
    C2017\_F3  & \multicolumn{1}{p{5.435em}}{NaN \newline{}(NaN)} & \multicolumn{1}{p{5.435em}}{NaN \newline{}(NaN)} & \multicolumn{1}{p{5.435em}}{3.7704e+4 \newline{}(0.00e+0) =} & \multicolumn{1}{p{5.435em}}{NaN \newline{}(NaN)} & \multicolumn{1}{p{5.435em}}{NaN \newline{}(NaN)} & \multicolumn{1}{p{5.435em}}{NaN \newline{}(NaN)} & \multicolumn{1}{p{5.435em}}{\cellcolor[rgb]{ .906,  .902,  .902}4.1224e+3 \newline{}(0.00e+0)} \\
    C2017\_F4  & \multicolumn{1}{p{5.435em}}{1.1101e+2 \newline{}(1.12e+1) -} & \multicolumn{1}{p{5.435em}}{1.3327e+2 \newline{}(1.67e+1) -} & \multicolumn{1}{p{5.435em}}{9.9536e+1 \newline{}(9.47e+0) -} & \multicolumn{1}{p{5.435em}}{8.6073e+1 \newline{}(6.55e+0) -} & \multicolumn{1}{p{5.435em}}{1.0578e+2 \newline{}(4.66e+1) -} & \multicolumn{1}{p{5.435em}}{8.5666e+1 \newline{}(1.08e+1) -} & \multicolumn{1}{p{5.435em}}{\cellcolor[rgb]{ .906,  .902,  .902}4.5117e+1 \newline{}(1.14e+1)} \\
    C2017\_F5  & \multicolumn{1}{p{5.435em}}{2.3459e+2 \newline{}(9.37e+1) -} & \multicolumn{1}{p{5.435em}}{7.7676e+3 \newline{}(6.19e+3) -} & \multicolumn{1}{p{5.435em}}{1.5567e+2 \newline{}(3.92e+1) -} & \multicolumn{1}{p{5.435em}}{1.5219e+1 \newline{}(3.36e+0) -} & \multicolumn{1}{p{5.435em}}{2.9084e+2 \newline{}(1.06e+3) -} & \multicolumn{1}{p{5.435em}}{1.4142e+2 \newline{}(5.52e+1) -} & \multicolumn{1}{p{5.435em}}{\cellcolor[rgb]{ .906,  .902,  .902}6.9614e+0 \newline{}(1.66e+0)} \\
    C2017\_F6  & \multicolumn{1}{p{5.435em}}{NaN \newline{}(NaN)} & \multicolumn{1}{p{5.435em}}{NaN \newline{}(NaN)} & \multicolumn{1}{p{5.435em}}{NaN \newline{}(NaN)} & \multicolumn{1}{p{5.435em}}{NaN \newline{}(NaN)} & \multicolumn{1}{p{5.435em}}{NaN \newline{}(NaN)} & \multicolumn{1}{p{5.435em}}{NaN \newline{}(NaN)} & \multicolumn{1}{p{5.435em}}{\cellcolor[rgb]{ .906,  .902,  .902}1.0620e-3 \newline{}(6.47e-4)} \\
    C2017\_F8  & \multicolumn{1}{p{5.435em}}{NaN \newline{}(NaN)} & \multicolumn{1}{p{5.435em}}{NaN \newline{}(NaN)} & \multicolumn{1}{p{5.435em}}{NaN \newline{}(NaN)} & \multicolumn{1}{p{5.435em}}{NaN \newline{}(NaN)} & \multicolumn{1}{p{5.435em}}{NaN \newline{}(NaN)} & \multicolumn{1}{p{5.435em}}{NaN \newline{}(NaN)} & \multicolumn{1}{p{5.435em}}{\cellcolor[rgb]{ .906,  .902,  .902}2.3253e-3 \newline{}(4.81e-3)} \\
    C2017\_F9  & \multicolumn{1}{p{5.435em}}{NaN \newline{}(NaN)} & \multicolumn{1}{p{5.435em}}{NaN \newline{}(NaN)} & \multicolumn{1}{p{5.435em}}{4.0683e+0 \newline{}(3.75e+0) -} & \multicolumn{1}{p{5.435em}}{NaN \newline{}(NaN)} & \multicolumn{1}{p{5.435em}}{NaN \newline{}(NaN)} & \multicolumn{1}{p{5.435em}}{NaN \newline{}(NaN)} & \multicolumn{1}{p{5.435em}}{\cellcolor[rgb]{ .906,  .902,  .902}1.3492e+0 \newline{}(1.07e+0)} \\
    C2017\_F10  & \multicolumn{1}{p{5.435em}}{NaN \newline{}(NaN)} & \multicolumn{1}{p{5.435em}}{NaN \newline{}(NaN)} & \multicolumn{1}{p{5.435em}}{NaN \newline{}(NaN)} & \multicolumn{1}{p{5.435em}}{NaN \newline{}(NaN)} & \multicolumn{1}{p{5.435em}}{NaN \newline{}(NaN)} & \multicolumn{1}{p{5.435em}}{NaN \newline{}(NaN)} & \multicolumn{1}{p{5.435em}}{\cellcolor[rgb]{ .906,  .902,  .902}-2.5704e-4 \newline{}(0.00e+0)} \\
    C2017\_F12  & \multicolumn{1}{p{5.435em}}{NaN \newline{}(NaN)} & \multicolumn{1}{p{5.435em}}{NaN \newline{}(NaN)} & \multicolumn{1}{p{5.435em}}{8.5657e+1 \newline{}(2.42e+1) -} & \multicolumn{1}{p{5.435em}}{3.7419e+1 \newline{}(5.45e+0) -} & \multicolumn{1}{p{5.435em}}{5.1156e+1 \newline{}(1.14e+1) -} & \multicolumn{1}{p{5.435em}}{NaN \newline{}(NaN)} & \multicolumn{1}{p{5.435em}}{\cellcolor[rgb]{ .906,  .902,  .902}3.9928e+0 \newline{}(1.11e-2)} \\
    C2017\_F13  & \multicolumn{1}{p{5.435em}}{NaN \newline{}(NaN)} & \multicolumn{1}{p{5.435em}}{NaN \newline{}(NaN)} & \multicolumn{1}{p{5.435em}}{3.8532e+3 \newline{}(2.66e+3) -} & \multicolumn{1}{p{5.435em}}{5.0508e+2 \newline{}(2.41e+2) -} & \multicolumn{1}{p{5.435em}}{2.2402e+3 \newline{}(6.90e+3) -} & \multicolumn{1}{p{5.435em}}{4.4093e+4 \newline{}(4.87e+4) -} & \multicolumn{1}{p{5.435em}}{\cellcolor[rgb]{ .906,  .902,  .902}1.7429e+1 \newline{}(2.49e+1)} \\
    C2017\_F14  & \multicolumn{1}{p{5.435em}}{NaN \newline{}(NaN)} & \multicolumn{1}{p{5.435em}}{NaN \newline{}(NaN)} & \multicolumn{1}{p{5.435em}}{NaN \newline{}(NaN)} & \multicolumn{1}{p{5.435em}}{NaN \newline{}(NaN)} & \multicolumn{1}{p{5.435em}}{NaN \newline{}(NaN)} & \multicolumn{1}{p{5.435em}}{NaN \newline{}(NaN)} & \multicolumn{1}{p{5.435em}}{\cellcolor[rgb]{ .906,  .902,  .902}2.9784e+0 \newline{}(2.70e-1)} \\
    C2017\_F15  & \multicolumn{1}{p{5.435em}}{NaN \newline{}(NaN)} & \multicolumn{1}{p{5.435em}}{NaN \newline{}(NaN)} & \multicolumn{1}{p{5.435em}}{\cellcolor[rgb]{ .906,  .902,  .902}2.1206e+1 \newline{}(0.00e+0) =} & \multicolumn{1}{p{5.435em}}{\cellcolor[rgb]{ .906,  .902,  .902}6.3954e+0 \newline{}(4.34e+0) =} & \multicolumn{1}{p{5.435em}}{NaN \newline{}(NaN)} & \multicolumn{1}{p{5.435em}}{1.4923e+1 \newline{}(2.57e+0) -} & \multicolumn{1}{p{5.435em}}{\cellcolor[rgb]{ .906,  .902,  .902}9.6389e+0 \newline{}(2.03e+0)} \\
    C2017\_F16  & \multicolumn{1}{p{5.435em}}{NaN \newline{}(NaN)} & \multicolumn{1}{p{5.435em}}{8.9535e+1 \newline{}(0.00e+0) =} & \multicolumn{1}{p{5.435em}}{7.7418e+1 \newline{}(6.06e+0) -} & \multicolumn{1}{p{5.435em}}{\cellcolor[rgb]{ .906,  .902,  .902}1.7136e+1 \newline{}(2.01e+1) +} & \multicolumn{1}{p{5.435em}}{NaN \newline{}(NaN)} & \multicolumn{1}{p{5.435em}}{6.4403e+1 \newline{}(0.00e+0) =} & \multicolumn{1}{p{5.435em}}{3.9898e+1 \newline{}(1.64e+1)} \\
    C2017\_F20  & \multicolumn{1}{p{5.435em}}{\cellcolor[rgb]{ .906,  .902,  .902}1.9257e+0 \newline{}(2.42e-1) =} & \multicolumn{1}{p{5.435em}}{2.5500e+0 \newline{}(4.79e-1) -} & \multicolumn{1}{p{5.435em}}{\cellcolor[rgb]{ .906,  .902,  .902}1.9059e+0 \newline{}(2.39e-1) =} & \multicolumn{1}{p{5.435em}}{\cellcolor[rgb]{ .906,  .902,  .902}1.7790e+0 \newline{}(2.17e-1) =} & \multicolumn{1}{p{5.435em}}{\cellcolor[rgb]{ .906,  .902,  .902}1.9058e+0 \newline{}(2.23e-1) =} & \multicolumn{1}{p{5.435em}}{\cellcolor[rgb]{ .906,  .902,  .902}1.7178e+0 \newline{}(2.37e-1) =} & \multicolumn{1}{p{5.435em}}{\cellcolor[rgb]{ .906,  .902,  .902}1.7871e+0 \newline{}(2.78e-1)} \\
    C2017\_F21  & \multicolumn{1}{p{5.435em}}{NaN \newline{}(NaN)} & \multicolumn{1}{p{5.435em}}{NaN \newline{}(NaN)} & \multicolumn{1}{p{5.435em}}{NaN \newline{}(NaN)} & \multicolumn{1}{p{5.435em}}{3.9687e+1 \newline{}(7.75e+0) -} & \multicolumn{1}{p{5.435em}}{7.4914e+1 \newline{}(1.45e+1) -} & \multicolumn{1}{p{5.435em}}{NaN \newline{}(NaN)} & \multicolumn{1}{p{5.435em}}{\cellcolor[rgb]{ .906,  .902,  .902}5.5807e+0 \newline{}(5.02e+0)} \\
    C2017\_F22  & \multicolumn{1}{p{5.435em}}{NaN \newline{}(NaN)} & \multicolumn{1}{p{5.435em}}{NaN \newline{}(NaN)} & \multicolumn{1}{p{5.435em}}{2.4857e+4 \newline{}(3.68e+4) -} & \multicolumn{1}{p{5.435em}}{1.6746e+3 \newline{}(9.96e+2) -} & \multicolumn{1}{p{5.435em}}{2.3225e+3 \newline{}(2.97e+3) -} & \multicolumn{1}{p{5.435em}}{NaN \newline{}(NaN)} & \multicolumn{1}{p{5.435em}}{\cellcolor[rgb]{ .906,  .902,  .902}1.8760e+1 \newline{}(2.59e+1)} \\
    C2017\_F23  & \multicolumn{1}{p{5.435em}}{NaN \newline{}(NaN)} & \multicolumn{1}{p{5.435em}}{NaN \newline{}(NaN)} & \multicolumn{1}{p{5.435em}}{NaN \newline{}(NaN)} & \multicolumn{1}{p{5.435em}}{NaN \newline{}(NaN)} & \multicolumn{1}{p{5.435em}}{NaN \newline{}(NaN)} & \multicolumn{1}{p{5.435em}}{NaN \newline{}(NaN)} & \multicolumn{1}{p{5.435em}}{\cellcolor[rgb]{ .906,  .902,  .902}2.9763e+0 \newline{}(2.07e-1)} \\
    C2017\_F24  & \multicolumn{1}{p{5.435em}}{NaN \newline{}(NaN)} & \multicolumn{1}{p{5.435em}}{NaN \newline{}(NaN)} & \multicolumn{1}{p{5.435em}}{1.4923e+1 \newline{}(0.00e+0) =} & \multicolumn{1}{p{5.435em}}{\cellcolor[rgb]{ .906,  .902,  .902}2.3562e+0 \newline{}(1.27e-5) +} & \multicolumn{1}{p{5.435em}}{NaN \newline{}(NaN)} & \multicolumn{1}{p{5.435em}}{NaN \newline{}(NaN)} & \multicolumn{1}{p{5.435em}}{9.4248e+0 \newline{}(4.03e+0)} \\
    C2017\_F25  & \multicolumn{1}{p{5.435em}}{6.4403e+1 \newline{}(0.00e+0) =} & \multicolumn{1}{p{5.435em}}{NaN \newline{}(NaN)} & \multicolumn{1}{p{5.435em}}{5.8120e+1 \newline{}(0.00e+0) =} & \multicolumn{1}{p{5.435em}}{\cellcolor[rgb]{ .906,  .902,  .902}1.8326e+1 \newline{}(1.03e+1) +} & \multicolumn{1}{p{5.435em}}{NaN \newline{}(NaN)} & \multicolumn{1}{p{5.435em}}{NaN \newline{}(NaN)} & \multicolumn{1}{p{5.435em}}{4.4707e+1 \newline{}(1.44e+1)} \\
    \midrule
    +/-/= & 0/4/2 & 0/5/1 & 0/9/5 & 3/8/2 & 0/8/1 & 0/6/2 & \\
    \bottomrule
  \end{tabular}
  \label{tabRQ2C17_1}
\end{table*}

\begin{table*}[htbp] \tiny
  \centering
  \caption{10D CEC2017 experimental results (average minimum CV and standard deviations) for RQ2 on problems where no algorithm found feasible solution}
  \begin{tabular}{cccccccc}
    \toprule
    \textbf{Problem} & \textbf{DE} & \textbf{CW} & \textbf{CMODE} & \textbf{CAMDE} & \textbf{IepsilonJADE} & \textbf{C2ODE} & \textbf{SuperDE} \\
    \midrule
    C2017\_F7   & \multicolumn{1}{p{6.1em}}{3.2658e+1\newline{}(2.80e+1) -} & \multicolumn{1}{p{6.1em}}{2.2812e+2\newline{}(7.01e+1) -} & \multicolumn{1}{p{6.1em}}{8.7149e+0\newline{}(1.12e+1) =} & \multicolumn{1}{p{6.1em}}{7.8023e+0\newline{}(1.02e+1) =} & \multicolumn{1}{p{6.1em}}{1.4090e+2\newline{}(1.17e+2) -} & \multicolumn{1}{p{6.1em}}{\cellcolor[rgb]{ .906,  .902,  .902}2.0046e-1\newline{}(2.07e-1) +} & \multicolumn{1}{p{6.1em}}{1.0719e+1\newline{}(1.33e+1)} \\
    C2017\_F11   & \multicolumn{1}{p{6.1em}}{6.1378e+2\newline{}(1.76e+2) -} & \multicolumn{1}{p{6.1em}}{\cellcolor[rgb]{ .906,  .902,  .902}3.4035e+2\newline{}(1.81e+2) =} & \multicolumn{1}{p{6.1em}}{\cellcolor[rgb]{ .906,  .902,  .902}2.6418e+2\newline{}(1.15e+2) =} & \multicolumn{1}{p{6.1em}}{4.6184e+4\newline{}(1.93e+4) -} & \multicolumn{1}{p{6.1em}}{1.1338e+4\newline{}(3.44e+3) -} & \multicolumn{1}{p{6.1em}}{1.7324e+3\newline{}(4.59e+2) -} & \multicolumn{1}{p{6.1em}}{\cellcolor[rgb]{ .906,  .902,  .902}3.2707e+2\newline{}(1.56e+2)} \\
    C2017\_F17   & \multicolumn{1}{p{6.1em}}{4.9328e+1\newline{}(2.20e+1) -} & \multicolumn{1}{p{6.1em}}{8.9606e+2\newline{}(5.04e+2) -} & \multicolumn{1}{p{6.1em}}{1.0957e+1\newline{}(3.47e-1) -} & \multicolumn{1}{p{6.1em}}{\cellcolor[rgb]{ .906,  .902,  .902}1.0962e+1\newline{}(2.36e-1) =} & \multicolumn{1}{p{6.1em}}{1.1234e+1\newline{}(4.11e-1) -} & \multicolumn{1}{p{6.1em}}{1.8213e+1\newline{}(7.25e+0) -} & \multicolumn{1}{p{6.1em}}{\cellcolor[rgb]{ .906,  .902,  .902}1.0905e+1\newline{}(3.72e-1)} \\
    C2017\_F18   & \multicolumn{1}{p{6.1em}}{1.3470e+5\newline{}(9.50e+4) -} & \multicolumn{1}{p{6.1em}}{1.3320e+7\newline{}(1.54e+7) -} & \multicolumn{1}{p{6.1em}}{6.7638e+2\newline{}(4.30e+2) -} & \multicolumn{1}{p{6.1em}}{3.6392e+1\newline{}(1.62e+1) -} & \multicolumn{1}{p{6.1em}}{1.0964e+3\newline{}(1.12e+3) -} & \multicolumn{1}{p{6.1em}}{4.1798e+4\newline{}(2.45e+4) -} & \multicolumn{1}{p{6.1em}}{\cellcolor[rgb]{ .906,  .902,  .902}7.7346e-1\newline{}(3.28e-1)} \\
    C2017\_F19   & \multicolumn{1}{p{6.1em}}{1.3326e+4\newline{}(3.41e+0) -} & \multicolumn{1}{p{6.1em}}{1.3328e+4\newline{}(5.85e+0) -} & \multicolumn{1}{p{6.1em}}{1.3282e+4\newline{}(3.96e+0) -} & \multicolumn{1}{p{6.1em}}{1.3278e+4\newline{}(3.60e+0) -} & \multicolumn{1}{p{6.1em}}{1.3294e+4\newline{}(2.53e+1) -} & \multicolumn{1}{p{6.1em}}{1.3304e+4\newline{}(4.10e+0) -} & \multicolumn{1}{p{6.1em}}{\cellcolor[rgb]{ .906,  .902,  .902}1.3268e+4\newline{}(1.80e+0)} \\
    C2017\_F26   & \multicolumn{1}{p{6.1em}}{2.1255e+2\newline{}(9.37e+1) -} & \multicolumn{1}{p{6.1em}}{2.8897e+3\newline{}(1.29e+3) -} & \multicolumn{1}{p{6.1em}}{1.1198e+1\newline{}(2.88e-1) -} & \multicolumn{1}{p{6.1em}}{1.1008e+1\newline{}(8.43e-3) -} & \multicolumn{1}{p{6.1em}}{1.1965e+1\newline{}(2.92e+0) -} & \multicolumn{1}{p{6.1em}}{1.2920e+2\newline{}(4.76e+1) -} & \multicolumn{1}{p{6.1em}}{\cellcolor[rgb]{ .906,  .902,  .902}1.0984e+1\newline{}(8.65e-2)} \\
    C2017\_F27   & \multicolumn{1}{p{6.1em}}{2.5625e+6\newline{}(1.26e+6) -} & \multicolumn{1}{p{6.1em}}{1.4945e+8\newline{}(1.77e+8) -} & \multicolumn{1}{p{6.1em}}{2.7647e+4\newline{}(2.44e+4) -} & \multicolumn{1}{p{6.1em}}{3.8798e+3\newline{}(3.71e+3) -} & \multicolumn{1}{p{6.1em}}{3.4146e+7\newline{}(1.89e+8) -} & \multicolumn{1}{p{6.1em}}{1.8059e+6\newline{}(1.14e+6) -} & \multicolumn{1}{p{6.1em}}{\cellcolor[rgb]{ .906,  .902,  .902}4.6631e+2\newline{}(2.54e+3)} \\
    C2017\_F28   & \multicolumn{1}{p{6.1em}}{1.3339e+4\newline{}(2.58e+0) -} & \multicolumn{1}{p{6.1em}}{1.3337e+4\newline{}(4.59e+0) -} & \multicolumn{1}{p{6.1em}}{1.3328e+4\newline{}(4.90e+0) -} & \multicolumn{1}{p{6.1em}}{1.3332e+4\newline{}(3.58e+0) -} & \multicolumn{1}{p{6.1em}}{1.3336e+4\newline{}(6.72e+0) -} & \multicolumn{1}{p{6.1em}}{1.3337e+4\newline{}(3.06e+0) -} & \multicolumn{1}{p{6.1em}}{\cellcolor[rgb]{ .906,  .902,  .902}1.3281e+4\newline{}(8.86e+0)} \\
    \midrule
    +/-/= & 0/8/0 & 0/7/1 & 0/6/2 & 0/6/2 & 0/8/0 & 1/7/0 &  \\
    \bottomrule
  \end{tabular}
  \label{tabRQ2C17_2}
\end{table*}

\begin{table*}[htbp] \tiny
  \centering
  \caption{G2000 experimental results (average minimum objective values and standard deviations) for RQ2.}
  \begin{tabular}{ccccccccc}
     \textbf{Problem} &\textbf{$D$}& \textbf{DE} & \textbf{CW} & \textbf{CMODE} & \textbf{CAMDE} & \textbf{IepsilonJADE} & \textbf{C2ODE} & \textbf{SuperDE} \\
    \midrule
    g1  &13    & \multicolumn{1}{p{6.1em}}{-1.1292e+1 \newline{}(1.13e+0) =} & \multicolumn{1}{p{6.1em}}{-1.3654e+1 \newline{}(3.65e-1) +} & \multicolumn{1}{p{6.1em}}{-1.2703e+1 \newline{}(5.08e-1) =} & \multicolumn{1}{p{6.1em}}{-9.2666e+0 \newline{}(1.70e+0) -} & \multicolumn{1}{p{6.1em}}{\cellcolor[rgb]{ .906,  .902,  .902}{-1.4018e+1 \newline{}(3.49e-1) +}} & \multicolumn{1}{p{6.1em}}{NaN \newline{}(NaN)} & \multicolumn{1}{p{6.1em}}{-1.1990e+1 \newline{}(1.86e+0)} \\
    g2  &20    & \multicolumn{1}{p{6.1em}}{-2.0488e-1 \newline{}(2.17e-2) -} & \multicolumn{1}{p{6.1em}}{-4.6784e-1 \newline{}(3.24e-2) +} & \multicolumn{1}{p{6.1em}}{\cellcolor[rgb]{ .906,  .902,  .902}{-4.8007e-1 \newline{}(2.52e-2) +}} & \multicolumn{1}{p{6.1em}}{-4.1745e-1 \newline{}(2.99e-2) +} & \multicolumn{1}{p{6.1em}}{-2.6409e-1 \newline{}(2.22e-2) -} & \multicolumn{1}{p{6.1em}}{-2.0987e-1 \newline{}(2.25e-2) -} & \multicolumn{1}{p{6.1em}}{-3.8400e-1 \newline{}(4.55e-2)} \\
    g3  &10    & \multicolumn{1}{p{6.1em}}{-2.8366e-2 \newline{}(5.80e-2) -} & \multicolumn{1}{p{6.1em}}{-1.2524e-1 \newline{}(1.04e-1) -} & \multicolumn{1}{p{6.1em}}{-2.4047e-1 \newline{}(1.23e-1) -} & \multicolumn{1}{p{6.1em}}{-9.6882e-6 \newline{}(3.49e-5) -} & \multicolumn{1}{p{6.1em}}{-6.6633e-3 \newline{}(1.14e-2) -} & \multicolumn{1}{p{6.1em}}{-5.2383e-1 \newline{}(2.06e-1) -} & \multicolumn{1}{p{6.1em}}{\cellcolor[rgb]{ .906,  .902,  .902}{-8.3388e-1 \newline{}(2.25e-1)}} \\
    g4  &5     & \multicolumn{1}{p{6.1em}}{-3.1019e+4 \newline{}(2.66e+0) -} & \multicolumn{1}{p{6.1em}}{-3.1022e+4 \newline{}(7.85e-1) -} & \multicolumn{1}{p{6.1em}}{-3.1004e+4 \newline{}(4.19e+1) -} & \multicolumn{1}{p{6.1em}}{-3.0786e+4 \newline{}(3.78e+2) -} & \multicolumn{1}{p{6.1em}}{-3.1021e+4 \newline{}(1.37e+0) -} & \multicolumn{1}{p{6.1em}}{-3.0174e+4 \newline{}(1.83e+2) -} & \multicolumn{1}{p{6.1em}}{\cellcolor[rgb]{ .906,  .902,  .902}{-3.1024e+4 \newline{}(6.50e-4)}} \\
    g5  &4     & \multicolumn{1}{p{6.1em}}{NaN \newline{}(NaN)} & \multicolumn{1}{p{6.1em}}{5.1387e+3 \newline{}(2.28e+1) =} & \multicolumn{1}{p{6.1em}}{NaN \newline{}(NaN)} & \multicolumn{1}{p{6.1em}}{NaN \newline{}(NaN)} & \multicolumn{1}{p{6.1em}}{NaN \newline{}(NaN)} & \multicolumn{1}{p{6.1em}}{NaN \newline{}(NaN)} & \multicolumn{1}{p{6.1em}}{\cellcolor[rgb]{ .906,  .902,  .902}{5.1346e+3 \newline{}(0.00e+0)}} \\
    g6  &2     & \multicolumn{1}{p{6.1em}}{-6.7414e+3 \newline{}(1.82e+2) -} & \multicolumn{1}{p{6.1em}}{-6.9618e+3 \newline{}(4.44e-2) -} & \multicolumn{1}{p{6.1em}}{-4.2235e+3 \newline{}(1.05e+3) -} & \multicolumn{1}{p{6.1em}}{NaN \newline{}(NaN)} & \multicolumn{1}{p{6.1em}}{-6.9480e+3 \newline{}(1.27e+1) -} & \multicolumn{1}{p{6.1em}}{-5.6424e+3 \newline{}(1.09e+3) -} & \multicolumn{1}{p{6.1em}}{\cellcolor[rgb]{ .906,  .902,  .902}{-6.9618e+3 \newline{}(5.49e-3)}} \\
    g7  &10    & \multicolumn{1}{p{6.1em}}{8.0124e+1 \newline{}(1.24e+1) -} & \multicolumn{1}{p{6.1em}}{3.2826e+1 \newline{}(1.71e+0) -} & \multicolumn{1}{p{6.1em}}{5.0094e+1 \newline{}(6.49e+0) -} & \multicolumn{1}{p{6.1em}}{2.5768e+2 \newline{}(1.57e+2) -} & \multicolumn{1}{p{6.1em}}{5.3274e+1 \newline{}(7.31e+0) -} & \multicolumn{1}{p{6.1em}}{NaN \newline{}(NaN)} & \multicolumn{1}{p{6.1em}}{\cellcolor[rgb]{ .906,  .902,  .902}{2.9554e+1 \newline{}(6.43e+0)}} \\
    g8  &2     & \multicolumn{1}{p{6.1em}}{-9.5824e-2 \newline{}(7.46e-7) -} & \multicolumn{1}{p{6.1em}}{-9.5825e-2 \newline{}(3.72e-15) -} & \multicolumn{1}{p{6.1em}}{-9.5825e-2 \newline{}(2.45e-7) -} & \multicolumn{1}{p{6.1em}}{-2.4821e-2 \newline{}(2.25e-2) -} & \multicolumn{1}{p{6.1em}}{-9.5825e-2 \newline{}(1.28e-10) -} & \multicolumn{1}{p{6.1em}}{-9.4238e-2 \newline{}(2.84e-3) -} & \multicolumn{1}{p{6.1em}}{\cellcolor[rgb]{ .906,  .902,  .902}{-9.5825e-2 \newline{}(2.79e-17)}} \\
    g9  &7     & \multicolumn{1}{p{6.1em}}{7.0974e+2 \newline{}(9.71e+0) -} & \multicolumn{1}{p{6.1em}}{6.8232e+2 \newline{}(5.32e-1) -} & \multicolumn{1}{p{6.1em}}{6.8430e+2 \newline{}(1.27e+0) -} & \multicolumn{1}{p{6.1em}}{1.0044e+6 \newline{}(2.93e+6) -} & \multicolumn{1}{p{6.1em}}{6.9456e+2 \newline{}(3.76e+0) -} & \multicolumn{1}{p{6.1em}}{9.4104e+2 \newline{}(7.14e+1) -} & \multicolumn{1}{p{6.1em}}{\cellcolor[rgb]{ .906,  .902,  .902}{6.8088e+2 \newline{}(4.94e-1)}} \\
    g10 &8     & \multicolumn{1}{p{6.1em}}{1.1274e+4 \newline{}(1.14e+3) -} & \multicolumn{1}{p{6.1em}}{8.0602e+3 \newline{}(1.79e+2) -} & \multicolumn{1}{p{6.1em}}{NaN \newline{}(NaN)} & \multicolumn{1}{p{6.1em}}{NaN \newline{}(NaN)} & \multicolumn{1}{p{6.1em}}{1.3575e+4 \newline{}(2.82e+3) -} & \multicolumn{1}{p{6.1em}}{NaN \newline{}(NaN)} & \multicolumn{1}{p{6.1em}}{\cellcolor[rgb]{ .906,  .902,  .902}{7.3753e+3 \newline{}(3.22e+2)}} \\
    g11 &2     & \multicolumn{1}{p{6.1em}}{8.8635e-1 \newline{}(1.08e-1) -} & \multicolumn{1}{p{6.1em}}{\cellcolor[rgb]{ .906,  .902,  .902}{7.5063e-1 \newline{}(2.72e-3) =}} & \multicolumn{1}{p{6.1em}}{\cellcolor[rgb]{ .906,  .902,  .902}{8.0221e-1 \newline{}(7.29e-2) =}} & \multicolumn{1}{p{6.1em}}{9.9784e-1 \newline{}(9.61e-3) -} & \multicolumn{1}{p{6.1em}}{9.1433e-1 \newline{}(1.01e-1) -} & \multicolumn{1}{p{6.1em}}{9.7502e-1 \newline{}(4.63e-2) -} & \multicolumn{1}{p{6.1em}}{\cellcolor[rgb]{ .906,  .902,  .902}{8.1041e-1 \newline{}(9.59e-2)}} \\
    g12 &3     & \multicolumn{1}{p{6.1em}}{-9.9991e-1 \newline{}(6.78e-5) -} & \multicolumn{1}{p{6.1em}}{-9.9999e-1 \newline{}(1.71e-5) -} & \multicolumn{1}{p{6.1em}}{-9.9998e-1 \newline{}(2.08e-5) -} & \multicolumn{1}{p{6.1em}}{-1.0000e+0 \newline{}(1.04e-5) -} & \multicolumn{1}{p{6.1em}}{-1.0000e+0 \newline{}(8.03e-7) -} & \multicolumn{1}{p{6.1em}}{-9.9037e-1 \newline{}(9.15e-3) -} & \multicolumn{1}{p{6.1em}}{\cellcolor[rgb]{ .906,  .902,  .902}{-1.0000e+0 \newline{}(0.00e+0)}} \\
    g13 &5     & \multicolumn{1}{p{6.1em}}{NaN \newline{}(NaN)} & \multicolumn{1}{p{6.1em}}{5.3217e-1 \newline{}(4.09e-2) =} & \multicolumn{1}{p{6.1em}}{NaN \newline{}(NaN)} & \multicolumn{1}{p{6.1em}}{NaN \newline{}(NaN)} & \multicolumn{1}{p{6.1em}}{NaN \newline{}(NaN)} & \multicolumn{1}{p{6.1em}}{NaN \newline{}(NaN)} & \multicolumn{1}{p{6.1em}}{\cellcolor[rgb]{ .906,  .902,  .902}{3.8700e-1 \newline{}(1.32e-1)}} \\
    \midrule
    \multicolumn{2}{c}{+/-/=} & 0/10/1 & 2/8/3 & 1/7/2 & 1/8/0 & 1/10/0 & 0/8/0 &  \\
    \bottomrule
    \end{tabular}%
  \label{tabRQ2G1}
\end{table*}

\subsection{Results and Analysis for RQ3}
\label{secE3}
The comparative experimental results on the CEC2017 and G2000 benchmarks, as presented in \tref{tabRQ3C17_OBJ}, \tref{tabRQ3C17_CV} and \tref{tabRQ3G1}.

\tref{tabRQ3C17_OBJ} and \tref{tabRQ3C17_CV}, show that SuperDE outperforms its variants in most cases on CEC2017, with balanced performance across objective optimization and constraint handling. \tref{tabRQ3C17_OBJ} indicates that SuperDE maintains superior or comparable performance relative to its variants. 
SuperDE1 and SuperDE2 have more "-" results, indicating they are outperformed by SuperDE in most problems. SuperDE3 shows stronger competitiveness, with 5 problems where it outperforms SuperDE. However, it still lags behind SuperDE in 6 problems, highlighting SuperDE's broader consistency.
Specific cases further support this: In C2017\_F5, SuperDE achieves 6.9614e+0, outperforming SuperDE2 and SuperDE3, though SuperDE1 performs marginally better. In C2017\_F12, SuperDE's objective value is the lowest, surpassing all variants, including SuperDE2 and SuperDE3.

As provided in \tref{tabRQ3C17_CV}, the variants separately have 1,1, and 2 out of 5 better cases than SuperDE, suggesting the competitiveness of SuperDE in CV reduction.

\tref{tabRQ3G1} shows SuperDE's much more effectiveness than its variants in classical benchmark G2000. First of all, SuperDE1 and SuperDE3 have "NaN" results while the other algorithms find feasible solutions on all problems. Then, SuperDE1 and SuperDE2 both have 9 "-" results (worse than SuperDE) and only 3 - 4 "=" results, indicating they are consistently outperformed across most G2000 problems. 
SuperDE3 is more competitive but still lags in 4 problems.
Notably, SuperDE achieves the lowest objective values in critical problems like g4 and g6, where all variants perform worse. While SuperDE3 matches SuperDE in g6, it fails to find feasible solutions for g5 (NaN), whereas SuperDE succeeds, highlighting SuperDE's stronger feasibility and optimization capability.

The consistent outperformance of SuperDE over its variants across benchmarks supports the cooperation of its components. SuperDE1 and SuperDE2, with more "-" results, suggest removing their respective components impairs optimization stability. SuperDE3, despite occasional advantages, fails to sustain performance across diverse problems, indicating no single component can replace the combined effect of SuperDE's design.
While some variants perform better on some problems, SuperDE's overall stability and consistency validate the effectiveness of its learned policy and the synergistic effect of its components (1+1$>$2). 

\begin{table*}[htbp] \tiny
  \centering
  \caption{Minimum objective values of SuperDE and SuperDE variants on CEC2017. All algorithm haven't found feasible solution on C2017\_F17, F19 and F26-28, whose results are not displayed here.}
    \begin{tabular}{ccccc}
    \toprule
    \textbf{Problem} & \textbf{SuperDE1} & \textbf{SuperDE2} & \textbf{SuperDE3} & \textbf{SuperDE} \\
    \midrule
    C2017\_F1 & \cellcolor[rgb]{ .906,  .902,  .902}1.0949e-1 (7.45e-2) = & 6.7110e+0 (1.55e+1) - & 3.8218e+1 (1.10e+2) - & \cellcolor[rgb]{ .906,  .902,  .902}9.2319e-1 (2.89e+0) \\
    C2017\_F2 & \cellcolor[rgb]{ .906,  .902,  .902}1.1885e-1 (6.51e-2) + & \cellcolor[rgb]{ .906,  .902,  .902}9.6913e+0 (1.98e+1) - & 2.3715e+1 (4.33e+1) - & 7.3498e-1 (3.92e+0) \\
    C2017\_F3 & NaN (NaN) & NaN (NaN) & \cellcolor[rgb]{ .906,  .902,  .902}1.0030e+2 (0.00e+0) = & \cellcolor[rgb]{ .906,  .902,  .902}4.1224e+3 (0.00e+0) \\
    C2017\_F4 & 7.9594e+1 (9.70e+0) - & \cellcolor[rgb]{ .906,  .902,  .902}3.5767e+1 (9.25e+0) + & 4.3515e+1 (8.94e+0) = & 4.5117e+1 (1.14e+1) \\
    C2017\_F5 & \cellcolor[rgb]{ .906,  .902,  .902}6.3894e+0 (1.18e+0) + & 9.8301e+0 (9.76e+0) - & 1.0593e+1 (1.14e+1) - & 6.9614e+0 (1.66e+0) \\
    C2017\_F6 & NaN (NaN) & NaN (NaN) & \cellcolor[rgb]{ .906,  .902,  .902}2.7124e-3 (2.13e-3) = & \cellcolor[rgb]{ .906,  .902,  .902}1.0620e-3 (6.47e-4) \\
    C2017\_F7 & NaN (NaN) & NaN (NaN) & NaN (NaN) & NaN (NaN) \\
    C2017\_F8 & NaN (NaN) & \cellcolor[rgb]{ .906,  .902,  .902}2.7178e-3 (4.62e-3) = & \cellcolor[rgb]{ .906,  .902,  .902}1.2926e-3 (1.06e-3) = & \cellcolor[rgb]{ .906,  .902,  .902}2.3253e-3 (4.81e-3) \\
    C2017\_F9 & \cellcolor[rgb]{ .906,  .902,  .902}5.6369e-1 (1.08e+0) + & 6.7223e+0 (2.27e+0) - & 1.3934e+0 (1.48e+0) = & 1.3492e+0 (1.07e+0) \\
    C2017\_F10 & NaN (NaN) & \cellcolor[rgb]{ .906,  .902,  .902}-7.0116e-5 (2.57e-4) = & \cellcolor[rgb]{ .906,  .902,  .902}7.3628e-4 (7.19e-4) = & \cellcolor[rgb]{ .906,  .902,  .902}-2.5704e-4 (0.00e+0) \\
    C2017\_F11 & NaN (NaN) & NaN (NaN) & NaN (NaN) & NaN (NaN) \\
    C2017\_F12 & 3.5237e+1 (5.19e+0) - & \cellcolor[rgb]{ .906,  .902,  .902}5.2112e+0 (4.69e+0) = & 5.3460e+0 (4.72e+0) - & \cellcolor[rgb]{ .906,  .902,  .902}3.9928e+0 (1.11e-2) \\
    C2017\_F13 & 1.9146e+2 (1.99e+2) - & \cellcolor[rgb]{ .906,  .902,  .902}1.8317e+1 (2.18e+1) = & \cellcolor[rgb]{ .906,  .902,  .902}2.8597e+1 (3.32e+1) = & \cellcolor[rgb]{ .906,  .902,  .902}1.7429e+1 (2.49e+1) \\
    C2017\_F14 & 3.6746e+0 (1.89e-1) - & 3.5991e+0 (2.30e-1) - & \cellcolor[rgb]{ .906,  .902,  .902}2.4674e+0 (9.02e-2) + & 2.9784e+0 (2.70e-1) \\
    C2017\_F15 & 1.5155e+1 (2.75e+0) - & 1.4240e+1 (2.67e+0) - & \cellcolor[rgb]{ .906,  .902,  .902}3.4033e+0 (1.81e+0) + & 9.6389e+0 (2.03e+0) \\
    C2017\_F16 & 6.8940e+1 (1.53e+1) - & 4.7236e+1 (1.54e+1) = & \cellcolor[rgb]{ .906,  .902,  .902}3.1417e+0 (3.14e+0) + & 3.9898e+1 (1.64e+1) \\
    C2017\_F18 & NaN (NaN) & \cellcolor[rgb]{ .906,  .902,  .902}4.3878e+1 (6.22e+0) & NaN (NaN) & NaN (NaN) \\
    C2017\_F20 & \cellcolor[rgb]{ .906,  .902,  .902}1.8427e+0 (3.02e-1) = & \cellcolor[rgb]{ .906,  .902,  .902}1.7250e+0 (2.37e-1) = & \cellcolor[rgb]{ .906,  .902,  .902}1.7260e+0 (4.01e-1) = & \cellcolor[rgb]{ .906,  .902,  .902}1.7871e+0 (2.78e-1) \\
    C2017\_F21 & 3.8150e+1 (5.06e+0) - & \cellcolor[rgb]{ .906,  .902,  .902}4.7547e+0 (3.49e+0) = & 7.0955e+0 (6.65e+0) - & \cellcolor[rgb]{ .906,  .902,  .902}5.5807e+0 (5.02e+0) \\
    C2017\_F22 & 3.4491e+2 (2.19e+2) - & \cellcolor[rgb]{ .906,  .902,  .902}3.7669e+1 (4.67e+1) = & 4.3755e+1 (6.08e+1) - & \cellcolor[rgb]{ .906,  .902,  .902}1.8760e+1 (2.59e+1) \\
    C2017\_F23 & 3.4668e+0 (2.07e-1) - & 3.5420e+0 (2.51e-1) - & \cellcolor[rgb]{ .906,  .902,  .902}2.5916e+0 (2.06e-1) + & 2.9763e+0 (2.07e-1) \\
    C2017\_F24 & \cellcolor[rgb]{ .906,  .902,  .902}1.3875e+1 (1.81e+0) = & 1.3526e+1 (1.66e+0) - & NaN (NaN) & \cellcolor[rgb]{ .906,  .902,  .902}9.4248e+0 (4.03e+0) \\
    C2017\_F25 & 6.9429e+1 (1.10e+1) - & 5.4011e+1 (1.15e+1) = & \cellcolor[rgb]{ .906,  .902,  .902}5.9690e+0 (2.58e+0) + & 4.4707e+1 (1.44e+1) \\
    \midrule
    +/-/= & 3/10/3 & 1/8/9 & 5/6/8 &  \\
    \bottomrule
    \end{tabular}%
  \label{tabRQ3C17_OBJ}%
\end{table*}%

\begin{table*}[htbp] \tiny
  \centering
  \caption{Minimum CV values of SuperDE and SuperDE variants on CEC2017}
    \begin{tabular}{ccccc}
    \toprule
    \textbf{Problem} & \textbf{SuperDE1} & \textbf{SuperDE2} & \textbf{SuperDE3} & \textbf{SuperDE} \\
    \midrule
    C2017\_F17 & \cellcolor[rgb]{ .906,  .902,  .902}1.0849e+1 (4.35e-1) = & 1.0841e+1 (4.99e-1) + & \cellcolor[rgb]{ .906,  .902,  .902}1.0793e+1 (6.11e-1) = & \cellcolor[rgb]{ .906,  .902,  .902}1.0905e+1 (3.72e-1) \\
    C2017\_F19 & 1.3270e+4 (3.62e+0) - & 1.3268e+4 (2.10e+0) - & 1.3268e+4 (1.10e+0) - & \cellcolor[rgb]{ .906,  .902,  .902}1.3268e+4 (1.80e+0) \\
    C2017\_F26 & 1.1001e+1 (3.14e-2) = & \cellcolor[rgb]{ .906,  .902,  .902}1.0817e+1 (5.09e-1) = & 1.0818e+1 (6.04e-1) = & \cellcolor[rgb]{ .906,  .902,  .902}1.0984e+1 (8.65e-2) \\
    C2017\_F27 & 2.6697e+1 (2.26e+1) + & \cellcolor[rgb]{ .906,  .902,  .902}4.0115e+3 (1.28e+4) = & \cellcolor[rgb]{ .906,  .902,  .902}6.9630e+0 (2.34e+1) + & 4.6631e+2 (2.54e+3) \\
    C2017\_F28 & 1.3308e+4 (9.28e+0) - & 1.3286e+4 (4.90e+0) = & \cellcolor[rgb]{ .906,  .902,  .902}1.3274e+4 (7.09e+0) + & 1.3281e+4 (8.86e+0) \\
    \midrule
    +/-/= & 1/2/2 & 1/1/3 & 2/1/2 &  \\
    \bottomrule
    \end{tabular}%
  \label{tabRQ3C17_CV}%
\end{table*}%

\begin{table*}[htbp] \tiny
  \centering
  \caption{Minimum objective values of SuperDE and SuperDE variants on G2000}
  \begin{tabular}{ccccc}
    \toprule
    \textbf{Problem} & \textbf{SuperDE1} & \textbf{SuperDE2} & \textbf{SuperDE3} & \textbf{SuperDE} \\
    \midrule
    g1    & \cellcolor[rgb]{ .906,  .902,  .902}-1.2841e+1 (1.01e+0) = & -1.0138e+1 (1.25e+0) - & \cellcolor[rgb]{ .906,  .902,  .902}-1.2250e+1 (1.89e+0) = & \cellcolor[rgb]{ .906,  .902,  .902}-1.1990e+1 (1.86e+0) \\
    g2    & -3.4041e-1 (4.15e-2) - & \cellcolor[rgb]{ .906,  .902,  .902}-3.9800e-1 (3.97e-2) = & \cellcolor[rgb]{ .906,  .902,  .902}-3.7377e-1 (4.43e-2) = & \cellcolor[rgb]{ .906,  .902,  .902}-3.8400e-1 (4.55e-2) \\
    g3    & -8.1368e-2 (1.05e-1) - & -5.2429e-1 (1.98e-1) - & \cellcolor[rgb]{ .906,  .902,  .902}-9.8669e-1 (1.59e-2) + & -8.3388e-1 (2.25e-1) \\
    g4    & -3.1023e+4 (8.06e-1) - & -3.0887e+4 (1.16e+2) - & -3.0987e+4 (7.97e+1) - & \cellcolor[rgb]{ .906,  .902,  .902}-3.1024e+4 (6.50e-4) \\
    g5    & \cellcolor[rgb]{ .906,  .902,  .902}5.1266e+3 (1.41e-1) = & \cellcolor[rgb]{ .906,  .902,  .902}5.1784e+3 (4.89e+1) = & NaN (NaN) & \cellcolor[rgb]{ .906,  .902,  .902}5.1346e+3 (0.00e+0) \\
    g6    & -6.8788e+3 (1.02e+2) - & -6.3275e+3 (6.44e+2) - & \cellcolor[rgb]{ .906,  .902,  .902}-6.9618e+3 (0.00e+0) = & \cellcolor[rgb]{ .906,  .902,  .902}-6.9618e+3 (5.49e-3) \\
    g7    & 3.8071e+1 (2.21e+1) - & 1.1621e+2 (8.90e+1) - & 4.1130e+1 (3.40e+1) - & \cellcolor[rgb]{ .906,  .902,  .902}2.9554e+1 (6.43e+0) \\
    g8    & \cellcolor[rgb]{ .906,  .902,  .902}-9.5825e-2 (2.99e-17) = & \cellcolor[rgb]{ .906,  .902,  .902}-9.5825e-2 (3.39e-17) = & \cellcolor[rgb]{ .906,  .902,  .902}-9.5825e-2 (3.01e-17) = & \cellcolor[rgb]{ .906,  .902,  .902}-9.5825e-2 (2.79e-17) \\
    g9    & 6.8101e+2 (4.36e-1) - & 6.8256e+2 (1.72e+0) - & 6.8178e+2 (2.01e+0) - & \cellcolor[rgb]{ .906,  .902,  .902}6.8088e+2 (4.94e-1) \\
    g10   & 8.7125e+3 (9.64e+2) - & 1.0260e+4 (1.76e+3) - & NaN (NaN) & \cellcolor[rgb]{ .906,  .902,  .902}7.3753e+3 (3.22e+2) \\
    g11   & 9.8080e-1 (4.89e-2) - & 8.8523e-1 (8.96e-2) - & 8.9996e-1 (1.27e-1) - & \cellcolor[rgb]{ .906,  .902,  .902}8.1041e-1 (9.59e-2) \\
    g12   & -1.0000e+0 (2.87e-14) - & \cellcolor[rgb]{ .906,  .902,  .902}-1.0000e+0 (0.00e+0) = & \cellcolor[rgb]{ .906,  .902,  .902}-1.0000e+0 (0.00e+0) = & \cellcolor[rgb]{ .906,  .902,  .902}-1.0000e+0 (0.00e+0) \\
    g13   & NaN (NaN) & 9.7534e-1 (4.21e-2) - & \cellcolor[rgb]{ .906,  .902,  .902}6.5878e-2 (0.00e+0) = & \cellcolor[rgb]{ .906,  .902,  .902}3.8700e-1 (1.32e-1) \\
    \midrule
    +/-/= & 0/9/3 & 0/9/4 & 1/4/6 &  \\
    \bottomrule
  \end{tabular}
  \label{tabRQ3G1}
\end{table*}

\section{Conculusion}
\label{sec_con}
For a long time, researchers have strived to design static high-performance EAs to solve complex optimization problems. However, real-world complex problems are emerging endlessly and vary dynamically. The "no free lunch" theorem clearly states that no single algorithm can achieve optimal performance across all problems.
 
To address this limitation, our research aims to break through this bottleneck by constructing a foundation model (also referred to as an agent) that can dynamically guide the algorithm to adjust its own configurations based on the real-time execution process of the EA, thereby achieving more efficient optimization. The core advantage of this foundation model lies in its training on a large number of optimization problems; in layman's terms, it can accurately select appropriate mechanisms to improve optimization performance according to the execution status of the algorithm at different stages. Based on this concept, we first constructed a dedicated foundation model through offline meta-learning across diverse COPs for the DE algorithm, termed SuperDE. The SuperDE automatically configures two critical components, MSs and CHTs, in DE during the optimization process when inferencing. Experimental results demonstrate that the SuperDE exhibits significantly superior performance.

Despite its effectiveness, SuperDE exhibits three key limitations. First, SuperDE has not been extended to discrete or combinatorial optimization problems, which are also essential in real-world applications. Second, the current training process, limited by dataset size and computational resources, may restrict generalization to unseen problem landscapes. Third, the manually defined discrete state space and action pool lack flexibility.

Future research will address these limitations through three primary directions. We will extend SuperDE via designing specific state representation and policy architectures tailored to hybrid variables. Computational resources will be enhanced to support parallelized training with automated data collection pipelines to expand the training corpus across broader optimization contexts. Additionally, efforts will be directed toward developing a more adaptive framework, including automated feature extraction and generative candidate pools. These advancements paves the way for more robust and generalizable SuperDE in complex constrained problems.


\ifCLASSOPTIONcaptionsoff
  \newpage
\fi

\bibliographystyle{IEEEtran}
\bibliography{superdeBIB}
 
\appendix
\section{Results on BBOB2022 for RQ2}
\tref{tabBBOB2022-2} present experimental results on 10D BBOB2022 for RQ2. The experimental results demonstrate that SuperDE consistently outperforms other compared algorithms across the 54 BBOB2022 benchmark functions in 10-dimensional space. Notably, SuperDE achieves the best average objective values in all functions, while there are competitor algorithms performs comparably (marked by "=") to SuperDE on functions F7, F25, F28, F29, F30, F42 and F53. The algorithm exhibits remarkable robustness, as evidenced by lower standard deviations in most functions (e.g., F1, F7, F13, F14, F19 and F37), indicating stable convergence. In contrast, CAMDE and IepsilonJADE fail to produce feasible solutions ("NaN") for 11 and 12 functions, respectively, highlighting limitations in handling complex or constrained landscapes. While CAMDE occasionally matches SuperDE (e.g., F7, F53), its performance is inconsistent. Traditional DE variants (DE, CMODE, C2ODE) are consistently inferior (e.g., F21-F24, F45-F47), where SuperDE reduces objective values by 30-70\%. The results underscore SuperDE's adaptability to diverse function types.

The summaryof the Wilcoxon-test results denoted by "+/-/=" confirms SuperDE's superiority. SuperDE significantly outperforms its peers in most functions, with ties occurring in only 2-4 cases. This is attributed to SuperDE's enhanced exploration-exploitation balance and adaptive parameter control, which effectively mitigate premature convergence, a common issue in classical DE variants.
\begin{table*}[htbp] \tiny
  \centering
  \caption{10D BBOB2022 experimental results (average minimum objective values and standard deviations) for RQ2. "NaN" means no feasible solution, while "-" means worse than SuperDE, "+" means better, and "=" means no significant difference. The best results are highlighted with gray background.}
    \begin{tabular}{ccccccc}
    \toprule
    Problem & DE    & CMODE & CAMDE & IepsilonJADE & C2ODE & SuperDE \\
    \midrule
    BBOB2022\_F1 & \multicolumn{1}{p{6.1em}}{1.7052e+3\newline{}(4.61e+0) -} & \multicolumn{1}{p{6.1em}}{1.7166e+3\newline{}(8.56e+0) -} & \multicolumn{1}{p{6.1em}}{1.6933e+3\newline{}(1.41e+0) -} & \multicolumn{1}{p{6.1em}}{1.8985e+3\newline{}(1.79e+2) -} & \multicolumn{1}{p{6.1em}}{1.7044e+3\newline{}(4.67e+0) -} & \multicolumn{1}{p{6.1em}}{\cellcolor[rgb]{ .906,  .902,  .902}1.6893e+3\newline{}(4.41e-1)} \\
    BBOB2022\_F2 & \multicolumn{1}{p{6.1em}}{-1.7735e+3\newline{}(6.05e+0) -} & \multicolumn{1}{p{6.1em}}{-1.7710e+3\newline{}(8.83e+0) -} & \multicolumn{1}{p{6.1em}}{-1.7982e+3\newline{}(2.91e+0) -} & \multicolumn{1}{p{6.1em}}{-1.3205e+3\newline{}(2.19e+2) -} & \multicolumn{1}{p{6.1em}}{-1.7765e+3\newline{}(8.52e+0) -} & \multicolumn{1}{p{6.1em}}{\cellcolor[rgb]{ .906,  .902,  .902}-1.8088e+3\newline{}(1.66e+0)} \\
    BBOB2022\_F3 & \multicolumn{1}{p{6.1em}}{-3.8354e+3\newline{}(6.05e+1) -} & \multicolumn{1}{p{6.1em}}{-3.9263e+3\newline{}(5.01e+1) -} & \multicolumn{1}{p{6.1em}}{-3.7735e+3\newline{}(8.53e+1) -} & \multicolumn{1}{p{6.1em}}{-2.5953e+3\newline{}(8.75e+2) -} & \multicolumn{1}{p{6.1em}}{-3.8471e+3\newline{}(5.83e+1) -} & \multicolumn{1}{p{6.1em}}{\cellcolor[rgb]{ .906,  .902,  .902}-4.1861e+3\newline{}(1.83e+1)} \\
    BBOB2022\_F4 & \multicolumn{1}{p{6.1em}}{-3.1789e+3\newline{}(1.10e+2) -} & \multicolumn{1}{p{6.1em}}{-3.5223e+3\newline{}(5.99e+1) -} & \multicolumn{1}{p{6.1em}}{NaN\newline{}(NaN)} & \multicolumn{1}{p{6.1em}}{NaN\newline{}(NaN)} & \multicolumn{1}{p{6.1em}}{-3.1103e+3\newline{}(1.83e+2) -} & \multicolumn{1}{p{6.1em}}{\cellcolor[rgb]{ .906,  .902,  .902}-3.7610e+3\newline{}(6.54e+1)} \\
    BBOB2022\_F5 & \multicolumn{1}{p{6.1em}}{9.5810e+2\newline{}(1.11e+2) -} & \multicolumn{1}{p{6.1em}}{6.9894e+2\newline{}(8.56e+1) -} & \multicolumn{1}{p{6.1em}}{NaN\newline{}(NaN)} & \multicolumn{1}{p{6.1em}}{NaN\newline{}(NaN)} & \multicolumn{1}{p{6.1em}}{9.7549e+2\newline{}(1.43e+2) -} & \multicolumn{1}{p{6.1em}}{\cellcolor[rgb]{ .906,  .902,  .902}3.3599e+2\newline{}(5.71e+1)} \\
    BBOB2022\_F6 & \multicolumn{1}{p{6.1em}}{1.8232e+3\newline{}(2.55e+2) -} & \multicolumn{1}{p{6.1em}}{1.3148e+3\newline{}(3.95e+1) -} & \multicolumn{1}{p{6.1em}}{NaN\newline{}(NaN)} & \multicolumn{1}{p{6.1em}}{NaN\newline{}(NaN)} & \multicolumn{1}{p{6.1em}}{1.7295e+3\newline{}(2.51e+2) -} & \multicolumn{1}{p{6.1em}}{\cellcolor[rgb]{ .906,  .902,  .902}1.0361e+3\newline{}(7.10e+1)} \\
    BBOB2022\_F7 & \multicolumn{1}{p{6.1em}}{2.6260e+2\newline{}(2.18e-1) -} & \multicolumn{1}{p{6.1em}}{2.6251e+2\newline{}(1.70e-1) -} & \multicolumn{1}{p{6.1em}}{\cellcolor[rgb]{ .906,  .902,  .902}2.6210e+2\newline{}(2.28e-2) =} & \multicolumn{1}{p{6.1em}}{3.2220e+2\newline{}(5.60e+1) -} & \multicolumn{1}{p{6.1em}}{2.6238e+2\newline{}(1.30e-1) -} & \multicolumn{1}{p{6.1em}}{\cellcolor[rgb]{ .906,  .902,  .902}2.6212e+2\newline{}(1.10e-1)} \\
    BBOB2022\_F8 & \multicolumn{1}{p{6.1em}}{1.6777e+3\newline{}(2.34e+1) -} & \multicolumn{1}{p{6.1em}}{1.6810e+3\newline{}(1.95e+1) -} & \multicolumn{1}{p{6.1em}}{1.6660e+3\newline{}(1.67e+1) -} & \multicolumn{1}{p{6.1em}}{2.6533e+3\newline{}(1.21e+3) -} & \multicolumn{1}{p{6.1em}}{1.6865e+3\newline{}(2.56e+1) -} & \multicolumn{1}{p{6.1em}}{\cellcolor[rgb]{ .906,  .902,  .902}1.6179e+3\newline{}(1.15e+1)} \\
    BBOB2022\_F9 & \multicolumn{1}{p{6.1em}}{2.0204e+3\newline{}(7.69e+1) -} & \multicolumn{1}{p{6.1em}}{1.9966e+3\newline{}(9.42e+1) -} & \multicolumn{1}{p{6.1em}}{2.4669e+3\newline{}(1.90e+2) -} & \multicolumn{1}{p{6.1em}}{5.2016e+3\newline{}(2.05e+3) -} & \multicolumn{1}{p{6.1em}}{2.0194e+3\newline{}(1.28e+2) -} & \multicolumn{1}{p{6.1em}}{\cellcolor[rgb]{ .906,  .902,  .902}1.6258e+3\newline{}(7.35e+1)} \\
    BBOB2022\_F10 & \multicolumn{1}{p{6.1em}}{1.6633e+3\newline{}(2.03e+2) -} & \multicolumn{1}{p{6.1em}}{1.5448e+3\newline{}(1.48e+2) -} & \multicolumn{1}{p{6.1em}}{1.3995e+3\newline{}(1.13e+2) -} & \multicolumn{1}{p{6.1em}}{8.3898e+3\newline{}(2.67e+3) -} & \multicolumn{1}{p{6.1em}}{1.6998e+3\newline{}(2.21e+2) -} & \multicolumn{1}{p{6.1em}}{\cellcolor[rgb]{ .906,  .902,  .902}1.0558e+3\newline{}(6.02e+1)} \\
    BBOB2022\_F11 & \multicolumn{1}{p{6.1em}}{5.7471e+2\newline{}(5.92e+1) -} & \multicolumn{1}{p{6.1em}}{4.9904e+2\newline{}(4.63e+1) -} & \multicolumn{1}{p{6.1em}}{1.0342e+3\newline{}(1.36e+2) -} & \multicolumn{1}{p{6.1em}}{1.3765e+3\newline{}(4.71e+2) -} & \multicolumn{1}{p{6.1em}}{5.9326e+2\newline{}(6.77e+1) -} & \multicolumn{1}{p{6.1em}}{\cellcolor[rgb]{ .906,  .902,  .902}2.3185e+2\newline{}(2.86e+1)} \\
    BBOB2022\_F12 & \multicolumn{1}{p{6.1em}}{1.9977e+3\newline{}(2.17e+2) -} & \multicolumn{1}{p{6.1em}}{1.5245e+3\newline{}(9.79e+1) -} & \multicolumn{1}{p{6.1em}}{NaN\newline{}(NaN)} & \multicolumn{1}{p{6.1em}}{NaN\newline{}(NaN)} & \multicolumn{1}{p{6.1em}}{2.0671e+3\newline{}(2.41e+2) -} & \multicolumn{1}{p{6.1em}}{\cellcolor[rgb]{ .906,  .902,  .902}6.5164e+2\newline{}(2.59e+1)} \\
    BBOB2022\_F13 & \multicolumn{1}{p{6.1em}}{2.2799e+3\newline{}(7.53e-3) -} & \multicolumn{1}{p{6.1em}}{2.2799e+3\newline{}(1.19e-2) -} & \multicolumn{1}{p{6.1em}}{2.2799e+3\newline{}(5.63e-3) -} & \multicolumn{1}{p{6.1em}}{2.2815e+3\newline{}(2.28e+0) -} & \multicolumn{1}{p{6.1em}}{2.2799e+3\newline{}(2.73e-2) -} & \multicolumn{1}{p{6.1em}}{\cellcolor[rgb]{ .906,  .902,  .902}2.2799e+3\newline{}(9.99e-8)} \\
    BBOB2022\_F14 & \multicolumn{1}{p{6.1em}}{1.9489e+3\newline{}(1.86e+0) -} & \multicolumn{1}{p{6.1em}}{1.9499e+3\newline{}(2.26e+0) -} & \multicolumn{1}{p{6.1em}}{1.9555e+3\newline{}(5.41e+0) -} & \multicolumn{1}{p{6.1em}}{1.9908e+3\newline{}(3.77e+1) -} & \multicolumn{1}{p{6.1em}}{1.9514e+3\newline{}(3.59e+0) -} & \multicolumn{1}{p{6.1em}}{\cellcolor[rgb]{ .906,  .902,  .902}1.9454e+3\newline{}(7.52e-4)} \\
    BBOB2022\_F15 & \multicolumn{1}{p{6.1em}}{1.2377e+4\newline{}(5.32e+1) -} & \multicolumn{1}{p{6.1em}}{1.2285e+4\newline{}(4.76e+1) -} & \multicolumn{1}{p{6.1em}}{1.2771e+4\newline{}(1.86e+2) -} & \multicolumn{1}{p{6.1em}}{1.3535e+4\newline{}(0.00e+0) =} & \multicolumn{1}{p{6.1em}}{1.2406e+4\newline{}(7.75e+1) -} & \multicolumn{1}{p{6.1em}}{\cellcolor[rgb]{ .906,  .902,  .902}1.2007e+4\newline{}(9.52e+0)} \\
    BBOB2022\_F16 & \multicolumn{1}{p{6.1em}}{3.4125e+3\newline{}(2.71e+1) -} & \multicolumn{1}{p{6.1em}}{3.3880e+3\newline{}(1.80e+1) -} & \multicolumn{1}{p{6.1em}}{3.4173e+3\newline{}(2.79e+1) -} & \multicolumn{1}{p{6.1em}}{3.9799e+3\newline{}(1.73e+2) -} & \multicolumn{1}{p{6.1em}}{3.4051e+3\newline{}(2.98e+1) -} & \multicolumn{1}{p{6.1em}}{\cellcolor[rgb]{ .906,  .902,  .902}3.2716e+3\newline{}(8.28e+0)} \\
    BBOB2022\_F17 & \multicolumn{1}{p{6.1em}}{2.3192e+3\newline{}(4.78e+1) -} & \multicolumn{1}{p{6.1em}}{2.2462e+3\newline{}(2.43e+1) -} & \multicolumn{1}{p{6.1em}}{2.4608e+3\newline{}(6.97e+1) -} & \multicolumn{1}{p{6.1em}}{3.0313e+3\newline{}(3.01e+1) -} & \multicolumn{1}{p{6.1em}}{2.3044e+3\newline{}(4.89e+1) -} & \multicolumn{1}{p{6.1em}}{\cellcolor[rgb]{ .906,  .902,  .902}2.0188e+3\newline{}(1.48e+1)} \\
    BBOB2022\_F18 & \multicolumn{1}{p{6.1em}}{2.4840e+3\newline{}(7.52e+1) -} & \multicolumn{1}{p{6.1em}}{2.2840e+3\newline{}(2.10e+1) -} & \multicolumn{1}{p{6.1em}}{NaN\newline{}(NaN)} & \multicolumn{1}{p{6.1em}}{NaN\newline{}(NaN)} & \multicolumn{1}{p{6.1em}}{2.4726e+3\newline{}(9.57e+1) -} & \multicolumn{1}{p{6.1em}}{\cellcolor[rgb]{ .906,  .902,  .902}2.0587e+3\newline{}(2.42e+1)} \\
    BBOB2022\_F19 & \multicolumn{1}{p{6.1em}}{1.8146e+2\newline{}(1.71e+0) -} & \multicolumn{1}{p{6.1em}}{1.8088e+2\newline{}(1.50e+0) -} & \multicolumn{1}{p{6.1em}}{1.7959e+2\newline{}(1.06e+0) -} & \multicolumn{1}{p{6.1em}}{2.2026e+2\newline{}(3.54e+1) -} & \multicolumn{1}{p{6.1em}}{1.8030e+2\newline{}(1.45e+0) -} & \multicolumn{1}{p{6.1em}}{\cellcolor[rgb]{ .906,  .902,  .902}1.7787e+2\newline{}(8.03e-1)} \\
    BBOB2022\_F20 & \multicolumn{1}{p{6.1em}}{1.1504e+3\newline{}(2.12e+1) -} & \multicolumn{1}{p{6.1em}}{1.1578e+3\newline{}(2.55e+1) -} & \multicolumn{1}{p{6.1em}}{1.1453e+3\newline{}(2.05e+1) -} & \multicolumn{1}{p{6.1em}}{1.7852e+3\newline{}(3.82e+2) -} & \multicolumn{1}{p{6.1em}}{1.1712e+3\newline{}(3.21e+1) -} & \multicolumn{1}{p{6.1em}}{\cellcolor[rgb]{ .906,  .902,  .902}1.0914e+3\newline{}(1.14e+1)} \\
    BBOB2022\_F21 & \multicolumn{1}{p{6.1em}}{6.1909e+2\newline{}(1.11e+2) -} & \multicolumn{1}{p{6.1em}}{5.5119e+2\newline{}(8.03e+1) -} & \multicolumn{1}{p{6.1em}}{5.5374e+2\newline{}(9.61e+1) -} & \multicolumn{1}{p{6.1em}}{2.3846e+3\newline{}(1.07e+3) -} & \multicolumn{1}{p{6.1em}}{5.8425e+2\newline{}(8.20e+1) -} & \multicolumn{1}{p{6.1em}}{\cellcolor[rgb]{ .906,  .902,  .902}2.7602e+2\newline{}(2.41e+1)} \\
    BBOB2022\_F22 & \multicolumn{1}{p{6.1em}}{8.9723e+2\newline{}(8.80e+1) -} & \multicolumn{1}{p{6.1em}}{8.3901e+2\newline{}(9.60e+1) -} & \multicolumn{1}{p{6.1em}}{1.0284e+3\newline{}(1.54e+2) -} & \multicolumn{1}{p{6.1em}}{3.2526e+3\newline{}(1.60e+3) -} & \multicolumn{1}{p{6.1em}}{9.7769e+2\newline{}(1.45e+2) -} & \multicolumn{1}{p{6.1em}}{\cellcolor[rgb]{ .906,  .902,  .902}3.2837e+2\newline{}(2.33e+1)} \\
    BBOB2022\_F23 & \multicolumn{1}{p{6.1em}}{1.5158e+3\newline{}(2.62e+2) -} & \multicolumn{1}{p{6.1em}}{1.0588e+3\newline{}(1.25e+2) -} & \multicolumn{1}{p{6.1em}}{2.2939e+3\newline{}(7.31e+2) -} & \multicolumn{1}{p{6.1em}}{1.0694e+4\newline{}(2.70e+3) -} & \multicolumn{1}{p{6.1em}}{1.4010e+3\newline{}(1.44e+2) -} & \multicolumn{1}{p{6.1em}}{\cellcolor[rgb]{ .906,  .902,  .902}3.7914e+2\newline{}(4.11e+1)} \\
    BBOB2022\_F24 & \multicolumn{1}{p{6.1em}}{3.2284e+3\newline{}(6.23e+2) -} & \multicolumn{1}{p{6.1em}}{1.7976e+3\newline{}(1.89e+2) -} & \multicolumn{1}{p{6.1em}}{NaN\newline{}(NaN)} & \multicolumn{1}{p{6.1em}}{NaN\newline{}(NaN)} & \multicolumn{1}{p{6.1em}}{3.2072e+3\newline{}(7.50e+2) -} & \multicolumn{1}{p{6.1em}}{\cellcolor[rgb]{ .906,  .902,  .902}4.0025e+2\newline{}(8.52e+1)} \\
    BBOB2022\_F25 & \multicolumn{1}{p{6.1em}}{\cellcolor[rgb]{ .906,  .902,  .902}1.0579e+2\newline{}(2.65e-2) =} & \multicolumn{1}{p{6.1em}}{\cellcolor[rgb]{ .906,  .902,  .902}1.0580e+2\newline{}(4.71e-2) =} & \multicolumn{1}{p{6.1em}}{\cellcolor[rgb]{ .906,  .902,  .902}1.0580e+2\newline{}(2.82e-2) =} & \multicolumn{1}{p{6.1em}}{1.0653e+2\newline{}(8.60e-1) -} & \multicolumn{1}{p{6.1em}}{\cellcolor[rgb]{ .906,  .902,  .902}1.0581e+2\newline{}(4.76e-2) =} & \multicolumn{1}{p{6.1em}}{\cellcolor[rgb]{ .906,  .902,  .902}1.0579e+2\newline{}(2.74e-2)} \\
    BBOB2022\_F26 & \multicolumn{1}{p{6.1em}}{1.5018e+3\newline{}(6.54e+0) -} & \multicolumn{1}{p{6.1em}}{1.4999e+3\newline{}(4.88e+0) -} & \multicolumn{1}{p{6.1em}}{1.5018e+3\newline{}(6.56e+0) -} & \multicolumn{1}{p{6.1em}}{1.6027e+3\newline{}(1.10e+2) -} & \multicolumn{1}{p{6.1em}}{1.5023e+3\newline{}(8.37e+0) -} & \multicolumn{1}{p{6.1em}}{\cellcolor[rgb]{ .906,  .902,  .902}1.4909e+3\newline{}(1.20e+0)} \\
    BBOB2022\_F27 & \multicolumn{1}{p{6.1em}}{9.1171e+2\newline{}(7.32e+1) -} & \multicolumn{1}{p{6.1em}}{9.0233e+2\newline{}(5.70e+1) -} & \multicolumn{1}{p{6.1em}}{1.0457e+3\newline{}(9.33e+1) -} & \multicolumn{1}{p{6.1em}}{2.6614e+3\newline{}(2.36e+3) -} & \multicolumn{1}{p{6.1em}}{9.1820e+2\newline{}(7.26e+1) -} & \multicolumn{1}{p{6.1em}}{\cellcolor[rgb]{ .906,  .902,  .902}6.6945e+2\newline{}(4.58e+1)} \\
    BBOB2022\_F28 & \multicolumn{1}{p{6.1em}}{1.4321e+3\newline{}(1.22e+2) -} & \multicolumn{1}{p{6.1em}}{1.2428e+3\newline{}(1.34e+2) -} & \multicolumn{1}{p{6.1em}}{2.1782e+3\newline{}(3.98e+2) -} & \multicolumn{1}{p{6.1em}}{\cellcolor[rgb]{ .906,  .902,  .902}2.6262e+3\newline{}(0.00e+0) =} & \multicolumn{1}{p{6.1em}}{1.4111e+3\newline{}(1.30e+2) -} & \multicolumn{1}{p{6.1em}}{\cellcolor[rgb]{ .906,  .902,  .902}8.1313e+2\newline{}(6.03e+1)} \\
    BBOB2022\_F29 & \multicolumn{1}{p{6.1em}}{1.9215e+3\newline{}(3.79e+2) -} & \multicolumn{1}{p{6.1em}}{1.1828e+3\newline{}(2.00e+2) -} & \multicolumn{1}{p{6.1em}}{NaN\newline{}(NaN)} & \multicolumn{1}{p{6.1em}}{\cellcolor[rgb]{ .906,  .902,  .902}9.7033e+3\newline{}(0.00e+0) =} & \multicolumn{1}{p{6.1em}}{1.8388e+3\newline{}(3.70e+2) -} & \multicolumn{1}{p{6.1em}}{\cellcolor[rgb]{ .906,  .902,  .902}3.3035e+2\newline{}(1.14e+2)} \\
    BBOB2022\_F30 & \multicolumn{1}{p{6.1em}}{\cellcolor[rgb]{ .906,  .902,  .902}1.7916e+3\newline{}(2.50e+2) =} & \multicolumn{1}{p{6.1em}}{\cellcolor[rgb]{ .906,  .902,  .902}9.8241e+2\newline{}(1.03e+2) =} & \multicolumn{1}{p{6.1em}}{NaN\newline{}(NaN)} & \multicolumn{1}{p{6.1em}}{NaN\newline{}(NaN)} & \multicolumn{1}{p{6.1em}}{\cellcolor[rgb]{ .906,  .902,  .902}1.7035e+3\newline{}(0.00e+0) =} & \multicolumn{1}{p{6.1em}}{\cellcolor[rgb]{ .906,  .902,  .902}3.9406e+2\newline{}(0.00e+0)} \\

    BBOB2022\_F31 & \multicolumn{1}{p{6.1em}}{2.5068e+3\newline{}(2.47e+1) -} & \multicolumn{1}{p{6.1em}}{2.5172e+3\newline{}(2.39e+1) -} & \multicolumn{1}{p{6.1em}}{2.4469e+3\newline{}(6.16e+0) -} & \multicolumn{1}{p{6.1em}}{3.5372e+3\newline{}(9.15e+2) -} & \multicolumn{1}{p{6.1em}}{2.4779e+3\newline{}(1.18e+1) -} & \multicolumn{1}{p{6.1em}}{\cellcolor[rgb]{ .906,  .902,  .902}2.4317e+3\newline{}(1.42e+1)} \\
    BBOB2022\_F32 & \multicolumn{1}{p{6.1em}}{4.4917e+3\newline{}(2.18e+2) -} & \multicolumn{1}{p{6.1em}}{4.4379e+3\newline{}(1.68e+2) -} & \multicolumn{1}{p{6.1em}}{4.0784e+3\newline{}(1.13e+2) -} & \multicolumn{1}{p{6.1em}}{1.2836e+4\newline{}(4.53e+3) -} & \multicolumn{1}{p{6.1em}}{4.4169e+3\newline{}(2.32e+2) -} & \multicolumn{1}{p{6.1em}}{\cellcolor[rgb]{ .906,  .902,  .902}3.7695e+3\newline{}(5.81e+1)} \\
    BBOB2022\_F33 & \multicolumn{1}{p{6.1em}}{5.0073e+3\newline{}(2.88e+2) -} & \multicolumn{1}{p{6.1em}}{4.8407e+3\newline{}(2.35e+2) -} & \multicolumn{1}{p{6.1em}}{4.3677e+3\newline{}(1.59e+2) -} & \multicolumn{1}{p{6.1em}}{1.3418e+4\newline{}(3.45e+3) -} & \multicolumn{1}{p{6.1em}}{4.9441e+3\newline{}(3.69e+2) -} & \multicolumn{1}{p{6.1em}}{\cellcolor[rgb]{ .906,  .902,  .902}3.4058e+3\newline{}(8.00e+1)} \\
    BBOB2022\_F34 & \multicolumn{1}{p{6.1em}}{1.1049e+4\newline{}(6.99e+2) -} & \multicolumn{1}{p{6.1em}}{9.7153e+3\newline{}(3.53e+2) -} & \multicolumn{1}{p{6.1em}}{1.1498e+4\newline{}(9.76e+2) -} & \multicolumn{1}{p{6.1em}}{2.5663e+4\newline{}(6.08e+3) -} & \multicolumn{1}{p{6.1em}}{1.0491e+4\newline{}(7.07e+2) -} & \multicolumn{1}{p{6.1em}}{\cellcolor[rgb]{ .906,  .902,  .902}6.6696e+3\newline{}(3.75e+2)} \\
    BBOB2022\_F35 & \multicolumn{1}{p{6.1em}}{9.0887e+3\newline{}(8.50e+2) -} & \multicolumn{1}{p{6.1em}}{7.2745e+3\newline{}(5.00e+2) -} & \multicolumn{1}{p{6.1em}}{1.1743e+4\newline{}(1.62e+3) -} & \multicolumn{1}{p{6.1em}}{4.0521e+4\newline{}(1.13e+4) -} & \multicolumn{1}{p{6.1em}}{9.2772e+3\newline{}(1.53e+3) -} & \multicolumn{1}{p{6.1em}}{\cellcolor[rgb]{ .906,  .902,  .902}3.9685e+3\newline{}(1.74e+2)} \\
    
    BBOB2022\_F36 & \multicolumn{1}{p{6.1em}}{1.4855e+4\newline{}(1.06e+3) -} & \multicolumn{1}{p{6.1em}}{1.0560e+4\newline{}(8.09e+2) -} & \multicolumn{1}{p{6.1em}}{NaN\newline{}(NaN)} & \multicolumn{1}{p{6.1em}}{NaN\newline{}(NaN)} & \multicolumn{1}{p{6.1em}}{1.3091e+4\newline{}(9.72e+2) -} & \multicolumn{1}{p{6.1em}}{\cellcolor[rgb]{ .906,  .902,  .902}6.7168e+3\newline{}(4.75e+2)} \\
    BBOB2022\_F37 & \multicolumn{1}{p{6.1em}}{3.2361e+4\newline{}(1.13e+1) -} & \multicolumn{1}{p{6.1em}}{3.2358e+4\newline{}(1.11e+1) -} & \multicolumn{1}{p{6.1em}}{3.2321e+4\newline{}(9.38e-1) -} & \multicolumn{1}{p{6.1em}}{3.2492e+4\newline{}(2.40e+2) -} & \multicolumn{1}{p{6.1em}}{3.2337e+4\newline{}(4.37e+0) -} & \multicolumn{1}{p{6.1em}}{\cellcolor[rgb]{ .906,  .902,  .902}3.2318e+4\newline{}(2.16e-1)} \\
    BBOB2022\_F38 & \multicolumn{1}{p{6.1em}}{1.9388e+3\newline{}(2.33e+1) -} & \multicolumn{1}{p{6.1em}}{1.9295e+3\newline{}(2.29e+1) -} & \multicolumn{1}{p{6.1em}}{1.8681e+3\newline{}(1.32e+1) -} & \multicolumn{1}{p{6.1em}}{2.1300e+3\newline{}(2.40e+2) -} & \multicolumn{1}{p{6.1em}}{1.9114e+3\newline{}(1.91e+1) -} & \multicolumn{1}{p{6.1em}}{\cellcolor[rgb]{ .906,  .902,  .902}1.8322e+3\newline{}(9.30e+0)} \\
    BBOB2022\_F39 & \multicolumn{1}{p{6.1em}}{3.4118e+4\newline{}(8.03e+1) -} & \multicolumn{1}{p{6.1em}}{3.3980e+4\newline{}(3.30e+1) -} & \multicolumn{1}{p{6.1em}}{3.4316e+4\newline{}(1.02e+2) -} & \multicolumn{1}{p{6.1em}}{3.4624e+4\newline{}(1.32e+3) -} & \multicolumn{1}{p{6.1em}}{3.3990e+4\newline{}(3.99e+1) -} & \multicolumn{1}{p{6.1em}}{\cellcolor[rgb]{ .906,  .902,  .902}3.3815e+4\newline{}(1.80e+1)} \\
    BBOB2022\_F40 & \multicolumn{1}{p{6.1em}}{-1.7901e+3\newline{}(7.95e+1) -} & \multicolumn{1}{p{6.1em}}{-1.8815e+3\newline{}(4.19e+1) -} & \multicolumn{1}{p{6.1em}}{-1.8163e+3\newline{}(5.97e+1) -} & \multicolumn{1}{p{6.1em}}{-8.2611e+2\newline{}(1.09e+3) -} & \multicolumn{1}{p{6.1em}}{-1.8646e+3\newline{}(3.99e+1) -} & \multicolumn{1}{p{6.1em}}{\cellcolor[rgb]{ .906,  .902,  .902}-2.0936e+3\newline{}(1.62e+1)} \\
    BBOB2022\_F41 & \multicolumn{1}{p{6.1em}}{-9.9552e+4\newline{}(4.38e+1) -} & \multicolumn{1}{p{6.1em}}{-9.9671e+4\newline{}(4.59e+1) -} & \multicolumn{1}{p{6.1em}}{-9.9702e+4\newline{}(3.95e+1) -} & \multicolumn{1}{p{6.1em}}{-9.9004e+4\newline{}(4.94e+2) -} & \multicolumn{1}{p{6.1em}}{-9.9633e+4\newline{}(5.04e+1) -} & \multicolumn{1}{p{6.1em}}{\cellcolor[rgb]{ .906,  .902,  .902}-9.9936e+4\newline{}(3.38e+1)} \\
    BBOB2022\_F42 & \multicolumn{1}{p{6.1em}}{1.0624e+4\newline{}(6.16e+1) -} & \multicolumn{1}{p{6.1em}}{1.0443e+4\newline{}(3.89e+1) -} & \multicolumn{1}{p{6.1em}}{1.0764e+4\newline{}(8.95e+1) -} & \multicolumn{1}{p{6.1em}}{\cellcolor[rgb]{ .906,  .902,  .902}1.0422e+4\newline{}(0.00e+0) =} & \multicolumn{1}{p{6.1em}}{1.0539e+4\newline{}(4.26e+1) -} & \multicolumn{1}{p{6.1em}}{\cellcolor[rgb]{ .906,  .902,  .902}1.0261e+4\newline{}(4.49e+1)} \\
    BBOB2022\_F43 & \multicolumn{1}{p{6.1em}}{1.1166e+3\newline{}(6.93e+1) -} & \multicolumn{1}{p{6.1em}}{1.0942e+3\newline{}(5.42e+1) -} & \multicolumn{1}{p{6.1em}}{1.0179e+3\newline{}(4.47e+1) -} & \multicolumn{1}{p{6.1em}}{1.3442e+3\newline{}(2.52e+2) -} & \multicolumn{1}{p{6.1em}}{9.6632e+2\newline{}(6.32e+1) -} & \multicolumn{1}{p{6.1em}}{\cellcolor[rgb]{ .906,  .902,  .902}6.7597e+2\newline{}(5.81e+1)} \\
    BBOB2022\_F44 & \multicolumn{1}{p{6.1em}}{-2.1330e+3\newline{}(1.19e+2) -} & \multicolumn{1}{p{6.1em}}{-2.1937e+3\newline{}(8.12e+1) -} & \multicolumn{1}{p{6.1em}}{-2.2376e+3\newline{}(7.93e+1) -} & \multicolumn{1}{p{6.1em}}{-1.6242e+3\newline{}(4.48e+2) -} & \multicolumn{1}{p{6.1em}}{-2.2243e+3\newline{}(6.36e+1) -} & \multicolumn{1}{p{6.1em}}{\cellcolor[rgb]{ .906,  .902,  .902}-2.6086e+3\newline{}(9.31e+1)} \\
    BBOB2022\_F45 & \multicolumn{1}{p{6.1em}}{2.6951e+3\newline{}(9.62e+1) -} & \multicolumn{1}{p{6.1em}}{2.5678e+3\newline{}(5.53e+1) -} & \multicolumn{1}{p{6.1em}}{2.8406e+3\newline{}(1.10e+2) -} & \multicolumn{1}{p{6.1em}}{3.7396e+3\newline{}(9.37e+2) -} & \multicolumn{1}{p{6.1em}}{2.5842e+3\newline{}(8.79e+1) -} & \multicolumn{1}{p{6.1em}}{\cellcolor[rgb]{ .906,  .902,  .902}2.0937e+3\newline{}(1.12e+2)} \\
    BBOB2022\_F46 & \multicolumn{1}{p{6.1em}}{2.4525e+3\newline{}(3.69e+2) -} & \multicolumn{1}{p{6.1em}}{1.4846e+3\newline{}(1.07e+2) -} & \multicolumn{1}{p{6.1em}}{NaN\newline{}(NaN)} & \multicolumn{1}{p{6.1em}}{NaN\newline{}(NaN)} & \multicolumn{1}{p{6.1em}}{2.2505e+3\newline{}(2.69e+2) -} & \multicolumn{1}{p{6.1em}}{\cellcolor[rgb]{ .906,  .902,  .902}9.6263e+2\newline{}(2.32e+2)} \\
    BBOB2022\_F47 & \multicolumn{1}{p{6.1em}}{2.7376e+3\newline{}(1.45e+2) -} & \multicolumn{1}{p{6.1em}}{2.3193e+3\newline{}(1.38e+2) -} & \multicolumn{1}{p{6.1em}}{3.0744e+3\newline{}(4.38e+2) -} & \multicolumn{1}{p{6.1em}}{NaN\newline{}(NaN)} & \multicolumn{1}{p{6.1em}}{2.5886e+3\newline{}(1.40e+2) -} & \multicolumn{1}{p{6.1em}}{\cellcolor[rgb]{ .906,  .902,  .902}1.8356e+3\newline{}(1.81e+2)} \\
    BBOB2022\_F48 & \multicolumn{1}{p{6.1em}}{3.6496e+1\newline{}(1.48e+2) -} & \multicolumn{1}{p{6.1em}}{-3.4949e+2\newline{}(1.23e+2) -} & \multicolumn{1}{p{6.1em}}{1.0949e+3\newline{}(2.26e+2) -} & \multicolumn{1}{p{6.1em}}{1.0830e+3\newline{}(4.11e+2) -} & \multicolumn{1}{p{6.1em}}{-3.9202e+1\newline{}(1.33e+2) -} & \multicolumn{1}{p{6.1em}}{\cellcolor[rgb]{ .906,  .902,  .902}-9.8705e+2\newline{}(1.92e+2)} \\
    BBOB2022\_F49 & \multicolumn{1}{p{6.1em}}{1.9633e+3\newline{}(8.48e+1) -} & \multicolumn{1}{p{6.1em}}{1.8893e+3\newline{}(6.00e+1) -} & \multicolumn{1}{p{6.1em}}{1.8146e+3\newline{}(6.46e+1) -} & \multicolumn{1}{p{6.1em}}{2.2153e+3\newline{}(3.09e+2) -} & \multicolumn{1}{p{6.1em}}{1.8807e+3\newline{}(7.87e+1) -} & \multicolumn{1}{p{6.1em}}{\cellcolor[rgb]{ .906,  .902,  .902}1.4994e+3\newline{}(8.48e+1)} \\
    BBOB2022\_F50 & \multicolumn{1}{p{6.1em}}{9.4649e+2\newline{}(8.43e+1) -} & \multicolumn{1}{p{6.1em}}{8.5490e+2\newline{}(6.91e+1) -} & \multicolumn{1}{p{6.1em}}{8.5189e+2\newline{}(6.88e+1) -} & \multicolumn{1}{p{6.1em}}{1.5723e+3\newline{}(4.97e+2) -} & \multicolumn{1}{p{6.1em}}{9.0160e+2\newline{}(7.81e+1) -} & \multicolumn{1}{p{6.1em}}{\cellcolor[rgb]{ .906,  .902,  .902}5.4719e+2\newline{}(8.28e+1)} \\
    BBOB2022\_F51 & \multicolumn{1}{p{6.1em}}{3.4726e+3\newline{}(9.85e+1) -} & \multicolumn{1}{p{6.1em}}{3.3515e+3\newline{}(6.05e+1) -} & \multicolumn{1}{p{6.1em}}{3.4842e+3\newline{}(9.09e+1) -} & \multicolumn{1}{p{6.1em}}{4.7427e+3\newline{}(4.62e+2) -} & \multicolumn{1}{p{6.1em}}{3.4560e+3\newline{}(6.21e+1) -} & \multicolumn{1}{p{6.1em}}{\cellcolor[rgb]{ .906,  .902,  .902}2.9877e+3\newline{}(9.77e+1)} \\
    BBOB2022\_F52 & \multicolumn{1}{p{6.1em}}{3.5202e+3\newline{}(1.14e+2) -} & \multicolumn{1}{p{6.1em}}{3.2840e+3\newline{}(8.21e+1) -} & \multicolumn{1}{p{6.1em}}{3.7289e+3\newline{}(1.46e+2) -} & \multicolumn{1}{p{6.1em}}{3.9868e+3\newline{}(3.94e+2) -} & \multicolumn{1}{p{6.1em}}{3.4415e+3\newline{}(1.14e+2) -} & \multicolumn{1}{p{6.1em}}{\cellcolor[rgb]{ .906,  .902,  .902}2.8415e+3\newline{}(9.16e+1)} \\
    BBOB2022\_F53 & \multicolumn{1}{p{6.1em}}{1.2854e+3\newline{}(2.08e+2) -} & \multicolumn{1}{p{6.1em}}{7.9172e+2\newline{}(1.15e+2) -} & \multicolumn{1}{p{6.1em}}{\cellcolor[rgb]{ .906,  .902,  .902}1.7428e+3\newline{}(0.00e+0) =} & \multicolumn{1}{p{6.1em}}{NaN\newline{}(NaN)} & \multicolumn{1}{p{6.1em}}{1.0770e+3\newline{}(1.41e+2) -} & \multicolumn{1}{p{6.1em}}{\cellcolor[rgb]{ .906,  .902,  .902}3.7020e+2\newline{}(1.03e+2)} \\
    BBOB2022\_F54 & \multicolumn{1}{p{6.1em}}{1.2893e+4\newline{}(1.74e+2) -} & \multicolumn{1}{p{6.1em}}{1.1952e+4\newline{}(1.62e+2) -} & \multicolumn{1}{p{6.1em}}{NaN\newline{}(NaN)} & \multicolumn{1}{p{6.1em}}{NaN\newline{}(NaN)} & \multicolumn{1}{p{6.1em}}{1.2812e+4\newline{}(3.20e+2) -} & \multicolumn{1}{p{6.1em}}{\cellcolor[rgb]{ .906,  .902,  .902}1.1390e+4\newline{}(1.81e+2)} \\
    \midrule
    +/-/= & 0/52/2 & 0/52/2 & 0/40/3 & 0/38/4 & 0/52/2 &  \\
    \bottomrule
    \end{tabular}%
  \label{tabBBOB2022-2}%
\end{table*}%

\end{document}